\documentclass[10pt,twocolumn,letterpaper]{article}

\usepackage{cvpr}              

\usepackage{graphicx}
\usepackage{amsmath}
\usepackage{amssymb}
\usepackage{booktabs}
\usepackage{mathtools,xparse}
\usepackage{placeins}
\usepackage[ruled,longend]{algorithm2e}
\usepackage{float}
\usepackage[accsupp]{axessibility} %

\DeclarePairedDelimiter{\norm}{\lVert}{\rVert}

\newcommand{\beginsupplement}{
  \setcounter{table}{0}  
  \renewcommand{\thetable}{S\arabic{table}} 
  \setcounter{figure}{0} 
  \renewcommand{\thefigure}{S\arabic{figure}}
}
\usepackage[pagebackref,breaklinks,colorlinks]{hyperref}

\usepackage[capitalize]{cleveref}
\crefname{section}{Sec.}{Secs.}
\Crefname{section}{Section}{Sections}
\Crefname{table}{Table}{Tables}
\crefname{table}{Tab.}{Tabs.}

\def\cvprPaperID{NA} 
\def\confName{CVPR}
\def\confYear{2023}

\begin{document}

\title{Out of Distribution Generalization via Interventional Style Transfer in Single-Cell Microscopy}

\author{Wolfgang M. Pernice$^*$\\
Columbia University\\
{\tt\small wp2181@cumc.columbia.edu}
\and
Michael Doron\\
Broad Institute\\
{\tt\small mdoron@broadinstitute.org}
\and
Alex Quach\\
MIT\\
{\tt\small aquach@mit.edu}
\and
Aditya Pratapa\\
Broad Institute\\
{\tt\small adyprat@pm.me}
\and
Sultan Kenjeyev\\
University College London\\
{\tt\small ksula0155@gmail.com}
\and
Nicholas De Veaux\\
New York University\\
{\tt\small nrdeveaux@gmail.com}
\and
Michio Hirano\\
Columbia University\\
{\tt\small mh29@cumc.columbia.edu}
\and
Juan C. Caicedo$^*$\\
Broad Institute\\
{\tt\small jcaicedo@broadinstitute.org}
}

\maketitle

\def\thefootnote{*}
\footnotetext{Co-corresponding authors}
\def\thefootnote{\arabic{footnote}}


\begin{abstract}
   \noindent Real-world deployment of computer vision systems, including in the discovery processes of biomedical research, requires causal representations that are invariant to contextual nuisances and generalize to new data. Leveraging the internal replicate structure of two novel single-cell fluorescent microscopy datasets, we propose generally applicable tests to assess the extent to which models learn causal representations across increasingly challenging levels of OOD-generalization. We show that despite seemingly strong performance, as assessed by other established metrics, both naive and contemporary baselines designed to ward against confounding, collapse on these tests. We introduce a new method, Interventional Style Transfer (IST), that substantially improves OOD generalization by generating interventional training distributions in which spurious correlations between biological causes and nuisances are mitigated. We publish our code\footnote{\href{https://github.com/Laboratory-for-Digital-Biology/InterventionalStyleTransfer}{https://github.com/Laboratory-for-Digital-Biology/IST}} and datasets\footnote{\href{10.5281/zenodo.7830240}{10.5281/zenodo.7830240}}.
\end{abstract}

\section{Introduction}

\label{sec:intro}
\noindent The ability to learn meaningful visual features from multiplexed microscopy images of cells and tissues promises to unlock cellular morphology as a powerful new single-cell omic with considerable potential to advance biomedical research \cite{mullard2019machine}. In turn, efforts are underway to collect fluorescent microscopy datasets that interrogate single-cell biology across hundreds of millions of cells and thousands of biological perturbations \cite{chandrasekaran2023jump, fay2023rxrx3}. To enable scientific discovery, computer vision models must learn representations that generalize to observations made in new observational environments (OEs) \cite{scholkopf2021toward, scholkopf2022causality}. Yet, vision systems are prone to learning spurious correlations between concepts of interest (e.g. objects) and contextual nuisances (e.g. background) \cite{mao2021generative}. This can yield biased representations that, although they may generalize well to hold-out sets that are independent and identically distributed (IID) with respect to the training data, collapse when tested on data that fall outside this distribution. For example, the performance of state-of-the-art (SOTA) vision models trained on the stereotypical views of objects in ImageNet, dramatically deteriorates when tested on ObjectNet images \cite{barbu2019objectnet}, which were collected with proactive interventions on several nuisance factors, such as background and object orientation (e.g. fallen-over chairs), that pose little challenge to humans. 

\begin{figure*}[!h]
  \centering
  \includegraphics[width=0.90\linewidth]{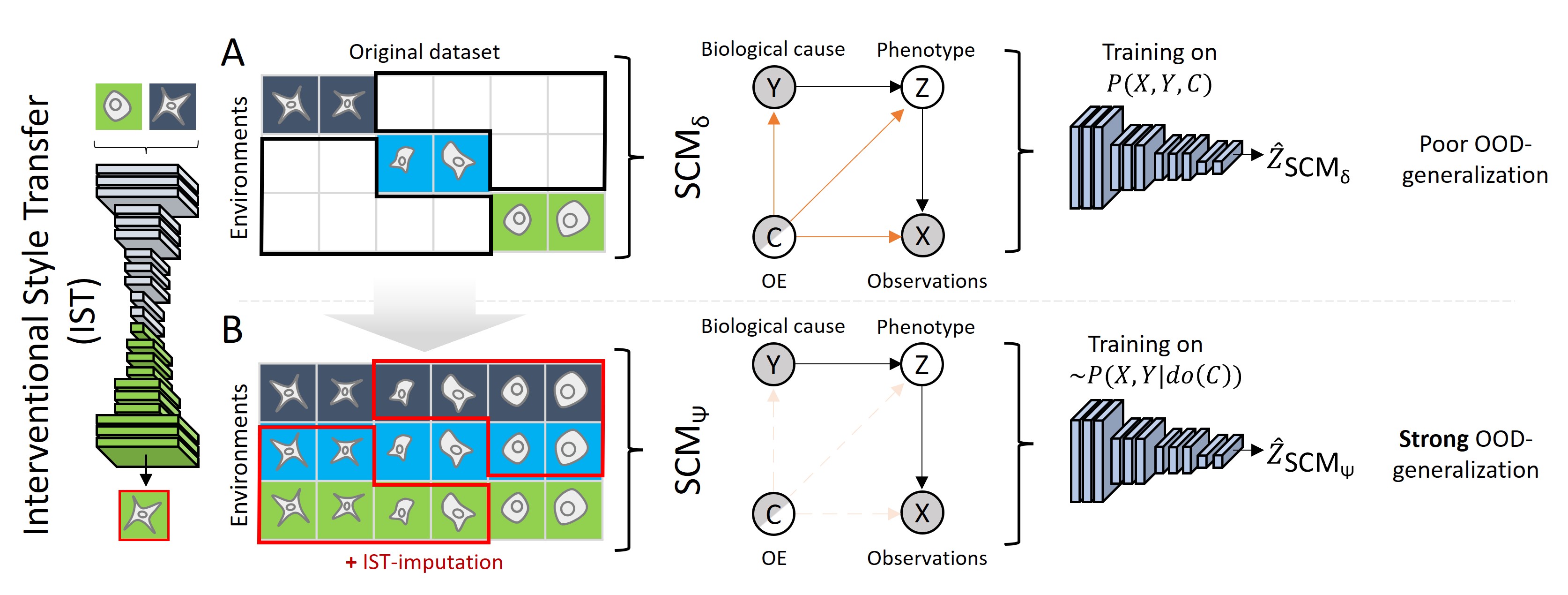}
  \vspace{-4mm}
  \caption{\small{(A) In most high-content datasets, not all conditions are observed in all OEs. Training distributions thus entail spurious correlations between OEs, observations and biological causes, yielding models that learn confounded representations according to $\mathbf{SCM}_{\delta}$. (B) We introduce an IST approach to impute images as if they had been collected in different OEs. By randomly permuting images across OEs, we yield an interventional distribution that removes spurious correlations with OEs allowing models to learn representations that are less biased and better capture the true causal structure according to $\mathbf{SCM}_{\psi}$.}}
  \label{fig:Figure_1_visual_abstract}
  \vspace{-4mm}
\end{figure*}

The same confounding influence that OEs exhibit in natural image datasets, manifests in biomedical datasets in the form of "batch-effects". Indeed, despite best efforts, technical variation between datasets collected in separate (experimental) batches cannot be perfectly controlled. Given the susceptibility of vision models to spurious correlations in natural image data outlined above, batch-effects present a major threat to meaningful biomedical applications of representation learning in fluorescent microscopy. 

In this paper we adopt the terminology of causal inference \cite{scholkopf2022causality, pearl2009causality} to study the (batch)effects of OEs as a confounder $\mathbf{C}$ to a general causal process that we suggest describes most datasets in biology. Our goal is to learn representations from such data, that model causal relationships, while remaining invariant to OEs. Importantly, we hold that the hypothesis that a given representation is causal (i.e. invariant over OEs and nuisances) cannot be falsified using IID hold-out data. Instead, we propose a rigorous testing regime based on generalization to OEs which are out-of-distribution (OOD) compared to the training data as a necessary characteristic and critical empirical measure of causal learning. To this end, and to foster progress towards causal representation learning in the field, we publicly release two real-world single-cell fluorescent microscopy datasets that exhibit internal replicate structures representative of most high-content imaging protocols (see below). We leverage this substructure to design realistic OOD generalization tasks. Surprisingly, we find that not only naive, but also SOTA post-processing and regularization baselines designed to mitigate batch-effects and improve generalization, fail when evaluated on these OOD tasks, despite in part excellent scores on IID hold-out sets and auxiliary metrics.

Given the ineffectiveness of existing methods on our OOD-task, we next consider intervening on the training distribution itself. Intuitively, if the training set contained observations balanced over all OEs, models should learn invariances to OEs and represent the right causal structure \cite{mao2021generative, pearl2009causality}. While collecting such dense datasets is not impossible (see e.g. \cite{schiff2022integrating}), many key applications require the assessment of very large numbers of conditions (e.g. over even modestly-sized drug libraries) that, for practical reasons, have to be collected in multiple batches. As such, most high-content imaging datasets are sparse, that is, sets of conditions are only observed in some OEs but not others (see Figs. \ref{fig:Figure_1_visual_abstract}, \ref{fig:Figure_2_Dataset_Overview}). Inspired by recent results on generative interventions to mitigate biases in natural image data \cite{mao2021generative}, we propose a new, light-weight method for \emph{Interventional Style Transfer} (IST) that generates effective interventions across an arbitrary number of OEs. To achieve this, we introduce architectural innovations and loss terms that prevent content hallucinations, which we find leads to failure of other style-transfer methods on our benchmark datasets. We then employ IST to yield a training distribution that mitigates OEs as confounders (Fig.\ref{fig:Figure_1_visual_abstract}) and show that models trained on it exhibit major improvements in OOD-generalization. 

As our main contributions, we (1) publish two new benchmark single-cell datasets with different degrees of sparsity in their replicate structure; we  (2) propose a rigorous OOD-generalization test regime that can be adopted across most experimental dataset; and (3) we introduce IST as the first method that achieves substantial improvements across increasingly challenging levels of OOD-generalization, as a starting-point for future work towards causal representation learning in microscopy and beyond. 

\section{Related Work}
\label{sec:Related Work}

\noindent \textbf{Data Augmentation:}
Data augmentations, such as blur, contrast, and rotations, are almost universally used in computer vision to yield more robust models \cite{lee2009advances, tian2020rethinking}. Both style-transfer \cite{geirhos2018imagenet, wang2017effectiveness} and adversarial training \cite{xie2020adversarial} have been employed in the pursuit of more complex augmentations. Our IST approach can be viewed as learning augmentations that imitate the effect of confounders. \vadjust{\vspace{3pt}}

\noindent \textbf{Generative Models:}
Generative models have been successfully employed on fluorescent microscopy and other biomedical data \cite{goldsborough2017cytogan, yang2021mol2image}. When OEs are unobserved, \cite{mao2021generative} show that generative models can be steered to produce noisy image manipulations on complex nuisances such as view point, that, when employed during training, improve OOD generalization. We employ the known replicate structure of our data to steer the generator directly. \vadjust{\vspace{3pt}}

\noindent \textbf{Domain Adaptation:}
Our approach builds on advances in domain-adaptation and style-transfer, developed to allow for the differential manipulating a style while preserving other content \cite{choi2020stargan, choi2017stargan, karras2020analyzing, lang2021explaining}. A major risk in applying style-transfer methods to scientific data is the inadvertent alteration of content (in our case phenotypic information). We hence design our IST approach to emphasizes content-preservation by discouraging major changes in pixel space. \vadjust{\vspace{-8pt}}

\noindent \textbf{Batch-effect correction:} How to mitigate batch-effects is an active field of study in biomedicine \cite{kang2021efficient}. Our work is closest to \cite{qian2020batch} who employ style-transfer to disentangle batch effects from biological features. IST features architectural improvements that prevent content alterations without the need for threshold- or segmentation-based regularization terms. IST also does not depend on assumptions about the nature of batch-effects (such as that they primarily manifest in first-order statistics \cite{zhou2021domain}), and achieves strong performance on challenging benchmarks without the need for additional post-hoc methods employed in \cite{qian2020batch}. \vadjust{\vspace{3pt}}

\noindent \textbf{Fairness:} A considerable body of work in visual recognition explores questions of fairness e.g. over demographic factors \cite{georgopoulos2021mitigating, yucer2020exploring}, including by means of style-transfer. Although discussions of causality are absent from these works, questions of fairness relate to our study on batch-effects as we seek to learn causal representations from biased data.  

\section{Causal Analysis}
\noindent Fig. \ref{fig:Figure_1_visual_abstract} presents a generalized structural causal model (SCM) \cite{pearl2009causality} that we assume as the basis of our work. We seek to reveal causal relationships between a set of conditions $\mathbf{Y}$ (e.g. disease categories) that may manifest cellular phenotypes $\mathbf{Z}$. To characterize $\mathbf{Z}$, we collect observations $\mathbf{X}$ using fluorescent microscopy. Observations are made in specific OEs $\mathbf{C}$ (i.e. batch, constituted by a specific well, plate, aliquot of reagents, etc.) that introduce technical variation to $\mathbf{X}$, and may further influence the biology of $\mathbf{Z}$, revealing it as a confounder \cite{scholkopf2022causality}. Importantly, in most datasets, not all conditions (we say, biological causes) are observed in all OEs. As such, the specific OE $\mathbf{c}$ also determines (the set of) biological causes $\mathbf{Y}$: e.g. two plates may contain different sets of conditions. Given training distributions $P(X,Y,C)$ in which biological causes are sparse over OEs, discriminative models learn spurious correlations between biological causes $\mathbf{Y}$ and confounding OEs $\mathbf{C}$ according to $\mathbf{SCM}_{\delta}$ resulting in biased representations $\mathbf{\hat{Z}_{{SCM}_{\delta}}}$ that generalize poorly to new OEs (Fig. \ref{fig:Figure_1_visual_abstract}A). 

A method that could produce faithful imputations of source images from one OE as if they had been collected in a different OE, could allow us to approximate the interventional distribution $P(X,Y|do(C))$ that would eliminate the backdoor paths emanating from $\mathbf{C}$, removing OEs as a confounder \cite{pearl2009causality}. Recent work on generative interventions for causal learning in natural images showed that even under noisy image manipulations, a classifier can learn better features for recognition in OOD data \cite{mao2021generative}. Inspired by their results, we propose an interventions-based approach that is compatible with experimental datasets. Importantly, in most natural image data-generation processes, observations cause labels, i.e. human experts label images according to what they see, and we train models to recapitulate this ability \cite{mao2021generative, recht2019imagenet}). Instead, in our datasets, $\mathbf{Y}$ represents conditions that we \emph{hypothesize} may cause observable cellular phenotypes. Our goal is then to approximate the conditional distribution $P(Y|X)$, that is to estimate the cause $\mathbf{Y}$ given noisy observations $\mathbf{X}$, whereby we hope to learn (discover) biologically meaningful representations of a priori unknown phenotypes $\mathbf{Z}$. Second, in contrast to natural images, where OEs are generally unobserved, our experimental data-acquisition protocols inherently document a rich ontology of processing steps that lead to any particular image (Fig. \ref{fig:Figure_2_Dataset_Overview}A). OEs are thus systematically tracked and feature rich metadata through which $\mathbf{C}$ is partially observed. In contrast to \cite{mao2021generative}, we can hence explicitly steer the data generation process learned by IST according to the known OE structure of our datasets. By using IST to intervene on $\mathbf{C}$, we seek to mitigate spurious correlations in the training distribution, yielding $\mathbf{\hat{Z}_{{SCM}_{\psi}}}$ and representations that generalize to OOD data (Fig. \ref{fig:Figure_1_visual_abstract}B).  

\begin{figure}
  \centering
  \includegraphics[width=0.90\linewidth]{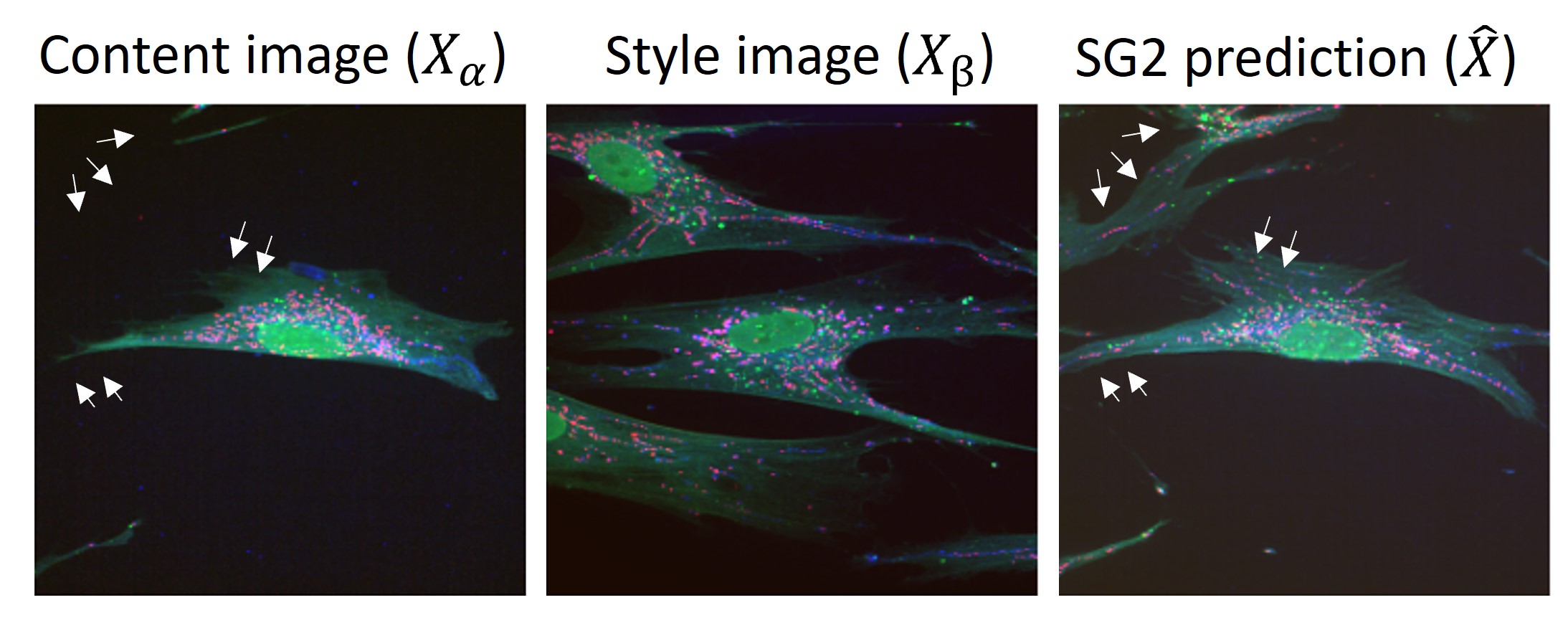}
  \vspace{-4mm}
  \caption{\small{StarGANv2 produces compelling images, but introduces both subtle and more obvious content alterations (white arrows)}}
  \label{fig:Figure_2_SG2}
  \vspace{-4mm}
\end{figure}

\section{Method}
\label{sec:Method}
 \noindent Advances in neural style transfer make it possible to perform image transformations that preserve spatial content while adjusting other feature statistics as desired \cite{karras2020analyzing, choi2020stargan, choi2017stargan, li2017universal}. In order to generate effective interventions on $\mathbf{C}$, a style-transfer model must learn to specifically transfer features related to OEs, while preserving phenotypic content. StarGANv2 \cite{choi2020stargan} introduced a framework to train a single encoder-decoder architecture capable of style-transfer between an arbitrary number of style-domains, such as demographic categories. We instead aim to steer our model to generate images in the "style" of specific OEs. Indeed, we find that StarGANv2, adapted to multichannel microscopy, produces visually compelling images. However, the outputs consistently feature both subtle and more obvious content hallucinations (Fig. \ref{fig:Figure_2_SG2}), suggesting that StarGANv2 fails to adequately preserve phenotypic content. 
 
 As a potential alternative to style-transfer in pixel space, \cite{zhou2021domain} propose MixStyle to pursue domain-generalization by mixing style-features of images from different OEs in the feature-maps of hidden-layers during training, with promising results. Remarkably, this avoids the need for a generator all together, making Mixstyle extremely lightweight. However \cite{zhou2021domain} assume that computing mean and standard-deviation of the feature-maps is sufficient to adequately capture style. While this may hold to some extent for natural images \cite{huang2017arbitrary}, there are no guarantees that this is true for batch-effects in microscopy data, and indeed, we find MixStyle offers little-to-no benefit on our task (see Sec. \ref{sec:Results}). 

To avoid these failure modes we design IST with several major and minor innovations. Specifically, to enforce content preservation, we introduce skip-connections between bottle-neck layers of our encoder and decoder, and introduce three complementary loss terms that discourage phenotypic alterations. Second, we find that, although first-order feature-map statistics are by themselves insufficient to describe OEs, they suffice as style-codes that, when injected into Adaptive Instance Normalization (AdaIN) layers \cite{huang2017arbitrary}, can be interpreted by our decoder to generate output images across an arbitrary number of OEs. This allows us to avoid all auxilliary networks required by StarGANv2 or comparable methods, rendering IST not only effective, but also computationally efficient, as detailed below. 

\subsection{Model Components}
\begin{figure*}
  \centering
  \includegraphics[width=0.90\linewidth]{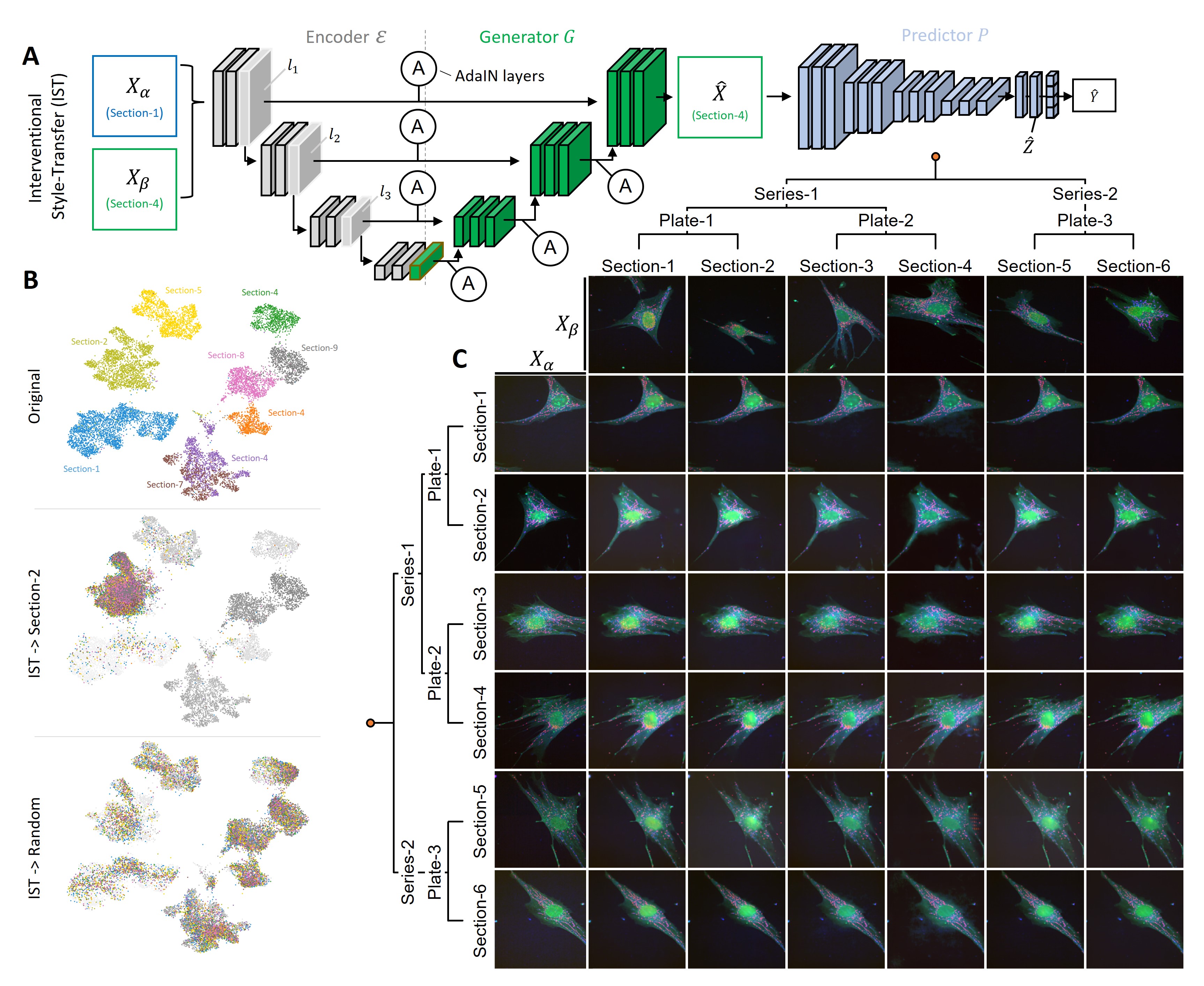}
  \vspace{-4mm}
  \caption{\small{A: Diagram of our IST-method. Given images $x_{\alpha}$ and $x_{\beta}$, encoder $\mathcal{E}$ (gray) extracts latent representations $u_{\alpha}$ and provides it to our generator (green) $G$. $\mathcal{E}$ further extracts style-codes $x_{\beta}$ and provides them to $G$ via AdaIN layers. We yield a prediction $\hat{x}$ that preserves the phenotypic content of $x_{\alpha}$ but inherits the OE of $x_{\beta}$. We train predictors $P$ (blue) on the resulting data distribution $\hat{X}$. B, C: UMAPs and output images illustrating the capacity of IST to project images into specific batches. When $x_{\beta}$ is sampled from a specific OE, output images fall onto their expected landmark in the UMAP space computed on the pretrained representations of $\mathcal{E}$ (see Sec. \ref{sec:Method}). When sampling $x_{\beta}$ fairly from all training OEs, the resulting distribution $\hat{X}$ is randomized over all OEs.
}}
  \label{fig:Figure_3_IST_MainFigure}
  \vspace{-4mm}
\end{figure*}

\noindent To train our IST-model, let $\mathbf{X}$ be the set of images, with associated environments $\mathbf{C}$ and cause labels $\mathbf{Y}$, respectively. Given an image $x \in \mathbf{X}$ observed in environment $c \in \mathbf{C}$, we seek to train a Generator $\mathcal{G}$ capable of producing image transformations $\hat{x}$ as if they come from other environments $\mathbf{C}$ (style) while preserving the original phenotypic content $z \in \mathbf{Z}$. To this end, we derive \emph{style codes} $v$ and train $\mathcal{G}$ to interpret them. Our framework consists of three main modules (see Fig. \ref{fig:Figure_3_IST_MainFigure}A): \vadjust{\vspace{3pt}}

\noindent \textbf{Encoder:} Given an image $x$, the encoder $\mathcal{E}$ derives the representation $u_x=\mathcal{E}(x)$, composed of multi-layer feature-sets $u_x^l$, with $l \in \{1...,L\}$ and $L$ the number of layers in the network. We implement $\mathcal{E}$ as a ResNet18 \cite{he2016deep} with instance normalization (IN) layers and a few modifications to facilitate skip connections. We pre-train $\mathcal{E}$ on an auxiliary multi-task objective of predicting $C$ and $Y$ given $X$. \vadjust{\vspace{3pt}}

\noindent \textbf{Generator:} Our generator predicts output images $\hat{x}=\mathcal{G}(u, v)$ given a feature set $u$ and an style code $v$. To promote the preservation of phenotypic content in the output images, we bias $\mathcal{G}$ against major changes in pixel space \cite{isola2017image}, by implementing $\mathcal{G}$ as a UNet-decoder with skip connections that concatenate feature-sets $u_x^l$ from the $l$-th corresponding feature-layer of the encoder, with $l \in \{1...,L\}$. We find that our choice improves both similarity between pairs $x$ and $\hat{x}$ as well as the realism of our output images. \vadjust{\vspace{3pt}}

\noindent \textbf{Critic:} Similar to \cite{choi2020stargan} we implement a critic $\mathcal{D}$ as a multi-task discriminator with $N_c$ output heads, where $N_c$ is the number of OEs. Each head $\mathcal{D}_c$ is trained as a binary classifier to distinguish real from fake images of their true or assigned OE $c$. To facilitate convergence, we initialize $\mathcal{D}$ with the weights of the pre-trained encoder $\mathcal{E}$ and fine-tune over the adversarial optimization process. \vadjust{\vspace{3pt}}

\noindent \textbf{Style codes:} To steer our generator, we compute image-specific style-codes. StarGANv2 employs a dedicated style-encoder to derive style-codes from latent distributions or input images \cite{choi2020stargan}. We find that effective style-codes can be computed directly from image features using our pre-trained, frozen encoder $\mathcal{E}$. Given the features $u_i=\mathcal{E}(x_i)$ of an image $x_i$, we compute:

\begin{equation} \label{eq:codes}
v_i = \left [ (\mu_{u_i^l}, \sigma_{u_i^l}) : l\in \{1...L\} \right ]
\end{equation}
where $\mu_{u_i^l}$ and $\sigma_{u_i^l}$ are the mean and standard deviation across the spatial domain of the feature maps of layer $l$. 

\begin{figure*}
  \centering
  \includegraphics[width=0.90\linewidth]{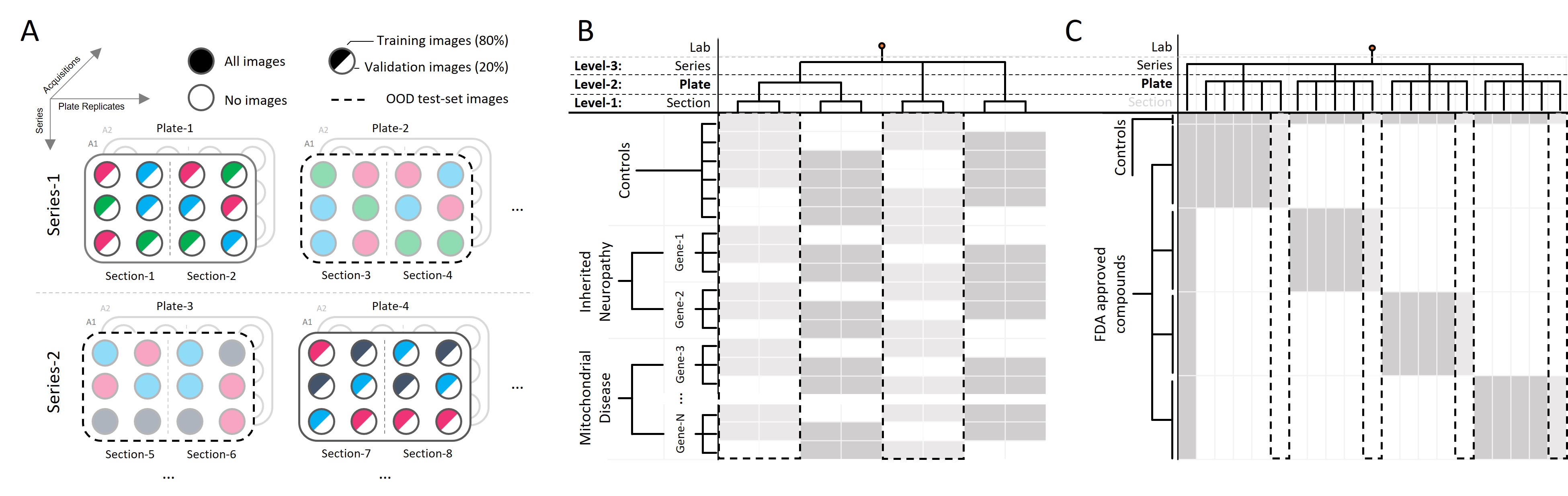}
  \vspace{-4mm}
  \caption{\small{A: diagrammatic illustration of a generalized data-acquisition process for high-content microscopy. Well colors indicate conditions (e.g. cell lines or perturbations) arrayed over multiwell plates with limited capacity. Datasets exhibit a nested replicate structures; a series constitutes a full experimental replicate including fresh cells and reagents, which may contain further replicates by plate, plate-section (acquired separately), and acquisitions, each constituting potentially meaningful OEs. This yields datasets of varying degrees of sparsity. B,C: schematic dataset substructure for GRID and LINCS-SC respectively. We define \emph{levels of generalization} according to increasingly distant relationships between OEs. The training and OOD-test setup for one fold of level-2 is indicated.}}
  \label{fig:Figure_2_Dataset_Overview}
  \vspace{-4mm}
\end{figure*} 

\subsection{Training the Generator} 
\noindent We train $\mathcal{G}$ to transform the appearance of an image from one OE to another. Following pre-training, we freeze the encoder $\mathcal{E}$ and use its weights to initialize the critic $\mathcal{D}$. The generator $\mathcal{G}$ is trained using SGD on pairs of triplets: $(x_\alpha,y_\alpha,c_\alpha)$ for content images and $(x_\beta,y_\beta,c_\beta)$ for style images. During training, we randomly sample content images balanced over $\mathbf{Y}$, and style images balanced over $\mathbf{C}$. In that way, content and style images are independently drawn to ensure samples with diverse phenotypic and technical variation respectively. During the forward pass, we first compute their feature sets $u_\alpha=\mathcal{E}(x_\alpha)$ and $u_\beta=\mathcal{E}(x_\beta)$ to then derive style codes $v_\alpha, v_\beta$. To intervene on the OE of the content image, we then inject $v_\beta$ using AdaIN-layers to predict the output $\hat{x}=\mathcal{G}(\mathcal{E}(x_\alpha),v_\beta)$. We minimize the following training objectives: \vadjust{\vspace{3pt}}

\noindent \textbf{Adversarial Loss:} Given a pair of content and style images $x_\alpha, x_\beta$, we compute style-codes $v_\alpha, v_\beta$ as described above. The generator is trained to produce realistic output images $\hat{x}=\mathcal{G}(\mathcal{E}(x_\alpha), v_\beta)$ with the following adversarial loss:
\begin{equation} \label{eq1}
\begin{split}
\mathcal{L}_{Adv}= & \mathbb{E}_{x,c}\left [ \log(\mathcal{D}_c(x) \right ] + \\
 & \mathbb{E}_{x,\tilde{x},\tilde{c}} \left [ \log(1 - \mathcal{D}_{\tilde{c}}(\mathcal{G}(\mathcal{E}(x), v_{\tilde{c}}))) \right ],
\end{split}
\end{equation}
where $\mathcal{D}_c(\cdot)$ is the head corresponding to OE $c$. $\mathcal{G}$ learns to use the style-codes $v_{\tilde{c}}$ to generate versions of $x$ as if observed in another environment $\tilde{c}$. \vadjust{\vspace{3pt}}

\noindent \textbf{Style Loss:} We further ensure effective intervention by applying a style-loss:
\begin{equation} \label{eq2}
\mathcal{L}_{Style} = \mathbb{E}_{\hat{x},c} \left [ 
\dfrac{1}{L} \sum_{l=1}^L \norm{ \text{Gram}(u^l_{x}) - \text{Gram}(u^l_{\hat{x}}) }_{1} 
\right ]
\end{equation}
where $\text{Gram}(\cdot)$ denotes the Gram matrix of features in the $l$-th layer of the encoder $\mathcal{E}$, used in style transfer to match the feature covariance of stylized images \cite{li2017universal, li2017demystifying}. \vadjust{\vspace{3pt}}

\noindent \textbf{Cycle-Consistency Loss:} To promote the preservation of phenotypic content of a source image $x_\alpha$ in the output $\hat{x}=\mathcal{G}(\mathcal{E}(x_\alpha), v_\beta)$, we apply a cycle-consistency loss \cite{zhu2017unpaired}:
\begin{equation} \label{eq3}
\mathcal{L}_{Cyc}=\mathbb{E}_{x,c}[\norm{x - \mathcal{G}(\mathcal{E}(\hat{x}), v_{c})}_{1}],
\end{equation}
where $v_c$ is the estimated style code of the original content image, i.e. we reconstruct $x_\alpha$ from $\hat{x}$. \vadjust{\vspace{3pt}}

\noindent \textbf{Content Loss:} We additionally constrain the absolute changes in pixel space between $x$ and $\hat{x}$ to prevent substantial loss or addition of phenotypic content (e.g. the hallucination of new cells or cellular components) by applying a content loss 
\begin{equation} \label{eq4}
\begin{split}
\mathcal{L}_{Cont}= \mathbb{E}_{\tilde{x},c} \left [ \norm{(\hat{x} - x)}_{1} + 
\dfrac{1}{L} \sum_{l=1}^L \norm{z^l_{\hat{x}} - z^l_{x}}_{1} \right ]
\end{split}
\end{equation}

\noindent \textbf{Class-matching Loss:} To further enforce that the generator preserves the phenotypic characteristics of input images, we implement a class-matching loss, defined as:
\begin{equation} \label{eq5}
\begin{split}
\mathcal{L}_{Cmatch} = \mathbb{E}_{\tilde{x}} \left [ -\sum_{y \in Y} \hat{y} ~
log \left ( \mathcal{E}_{cmt}(\hat{x})_y \right ) \right ],
\end{split}
\end{equation}
which is essentially the cross entropy loss of the cause predictions for the synthesized image with respect to the predictions for the real input image, according to the frozen encoder classifier $\mathcal{E}_{cls}$. Note that instead of using the actual cause label $y$, we use as target the prediction for the real image $\hat{y}=\mathcal{E}_{cls}(x_c)_y$. \vadjust{\vspace{3pt}}

\noindent \textbf{Full Objective:} We then optimize a min-max objective that trains the generator and critic in an adversarial fashion:
\begin{equation} \label{eq:full}
\begin{split}
\min_{\mathcal{G}} \max_{\mathcal{D}} = & \mathcal{L}_{Adv} + \lambda_{1}\mathcal{L}_{Style} + \\ 
& \lambda_{2}\mathcal{L}_{Cyc} + \lambda_{3}\mathcal{L}_{Cont} + \lambda_{4}\mathcal{L}_{Cmatch},
\end{split}
\end{equation}
where $\lambda_i \in \mathbb{R}$ are hyperparameters of the loss terms. 

\subsection{IST for causal learning}
\label{sec:IST for causal learning}
\noindent Once our IST-model is trained, we employ it to generate an interventional training distribution $P(\hat{X},Y|do(C))$, on which we in turn train a predictor network $P$ (Fig. \ref{fig:Figure_3_IST_MainFigure}A). To produce $\hat{X}$ during predictor training, content $x_\alpha$ and style $x_\beta$ images are sampled from the training distribution by pairing random causes with random OEs (both drawn uniformly) and passed through the (frozen) IST-model. This strategy breaks the spurious correlations between biological causes and OEs present in the original datasets. During testing, we also pass test images through our IST-model by randomly pairing them with a training image. This can be interpreted as bringing unseen images to familiar OEs for analysis, and we observed that IST-trained predictors perform better using this additional test-time correction.

\section{Experiments}
\label{sec:Experiments}
 \noindent To evaluate the merits of our IST approach in causal representation learning compared to relevant contemporary baselines, we conduct experiments on two novel  
 single-cell microscopy datasets that exhibit different degrees of sparsity and correlation between biological causes and OEs (Fig. \ref{fig:Figure_2_Dataset_Overview}B,D). Based on the known OE substructure in these dataset, we propose and empirically assess three increasingly challenging levels of OOD-generalization, by constructing hold-out sets according to a hierarchy of processing steps that separate them from the training data (Fig.\ref{fig:Figure_2_Dataset_Overview}).\footnote{Although we offer some details on biological causes for context, we do not interpret our results with respect to their biological implications; we focus on testing the generality of models across OEs.}

\subsection{Datasets}
\noindent \textbf{GRID:} We publish a subset of the \emph{Genetics of Rare Inherited Disease} (GRID) dataset, collected to discover latent disease-associated phenotypes in primary patient cells. The dataset contains 17,030 fluorescent microscopy images that reveal the organelle structure of primary dermal fibroblasts derived from 19 patients with 8 genetically confirmed inherited mitochondrial or neuromuscular diseases, and healthy controls (Fig.\ref{fig:Figure_2_Dataset_Overview}B). Data was acquired in multi-well plates with a hierarchical replicate structure: images were collected within the minimal OE of individual wells that contain cells of a specific cell-line. Images of the same cell-lines were collected in multiple (replicate) wells, organized into plate-\emph{sections}, \emph{plates}, and \emph{series} (Fig. \ref{fig:Figure_2_Dataset_Overview}A,B). Replicate wells across \emph{sections} (level-1) are seeded onto the same plate, during the same tissue culture session and derive from the same source cultures. Plate-level replicates (level-2) are separated by \emph{plate}, but share source cultures. Finally, \emph{series} (level-3) indicate full experimental replicates, starting with fresh thaws of cells. Critically, while sections contain identical sets of cell-lines, they only partially overlap between plates and series, yielding a sparse matrix of biological causes vs. OEs (Fig. \ref{fig:Figure_2_Dataset_Overview}B). \vadjust{\vspace{3pt}}

\noindent \textbf{LINCS-SC:} In contrast to GRID, the LINCS Cell-Profiling dataset was collected as a pharmacological perturbation study, including 1,327 clinically relevant compounds \cite{corsello2017drug}, using a \emph{single} A549 lung cancer cell line \cite{way2022morphology}. Cells were stained according to the Cell-Painting protocol \cite{bray2016cell} and imaged at lower magnification, such that the resulting images contain many cells. The LINCS single-cell (LINCS-SC) dataset, contains a subset of 101 compounds with strong morphological effects as judged by prior analyses \cite{moshkov2022learning}. Single-cell images were derived by segmenting source images with Cell Profiler \cite{mcquin2018cellprofiler} for a total of 200,000 images. In contrast to GRID, LINCS plates contain no sections. Moreover, LINCS plate- and series-level replicates are structured according to 25 unique plate-maps that host exclusive perturbations: with the exception of controls, there is no overlap between compounds across plate-maps. Consequently, the data-matrix is almost perfectly sparse between plate-maps (Fig. \ref{fig:Figure_2_Dataset_Overview}C). Finally, in LINCS, only one series contains all plate-maps (i.e. treatments), but without plate-level replicates, while four additional series contain exclusive subsets of plate-maps, each replicated 5 times (Fig. \ref{fig:Figure_2_Dataset_Overview}C).

\subsection{Baselines}
\noindent We seek to train predictors $P$ such that they generalizes to unseen OEs. We compare IST to strong domain-specific and more general baselines that collectively represent three major categories: post-hoc correction in feature space, regularization during training, and interventions on the training distribution. For all experiments, we use the same set of pre-processing steps and augmentations. As a naive baseline, we randomly initialize a predictor $P$ as a ResNet18 network (using IN layers) attached to a linear classification head and train it to predict biological causes $\mathbf{Y}$ from $\mathbf{\hat{X}}$. We implement other methods via minimal necessary deviations: \vadjust{\vspace{-8pt}}

\noindent \textbf{Symphony:} Symphony (SYM) is a state-of-the-art batch-effect correction method developed for single-cell RNA-sequencing (scRNAseq) datasets \cite{kang2021efficient}. Symphony extends a previous method, Harmony \cite{korsunsky2019fast}, which learns linear corrections over labeled nuisances. In contrast to Harmony, Symphony allows for inference on unseen datasets. We fit Symphony on training-set features $\hat{Z}$ extracted from our naive baseline. We set the \emph{topn} hyperparameter equal to our feature-dimension and empirically tune others. \vadjust{\vspace{3pt}}

\begin{figure}
  \centering
  \includegraphics[width=0.90\linewidth]{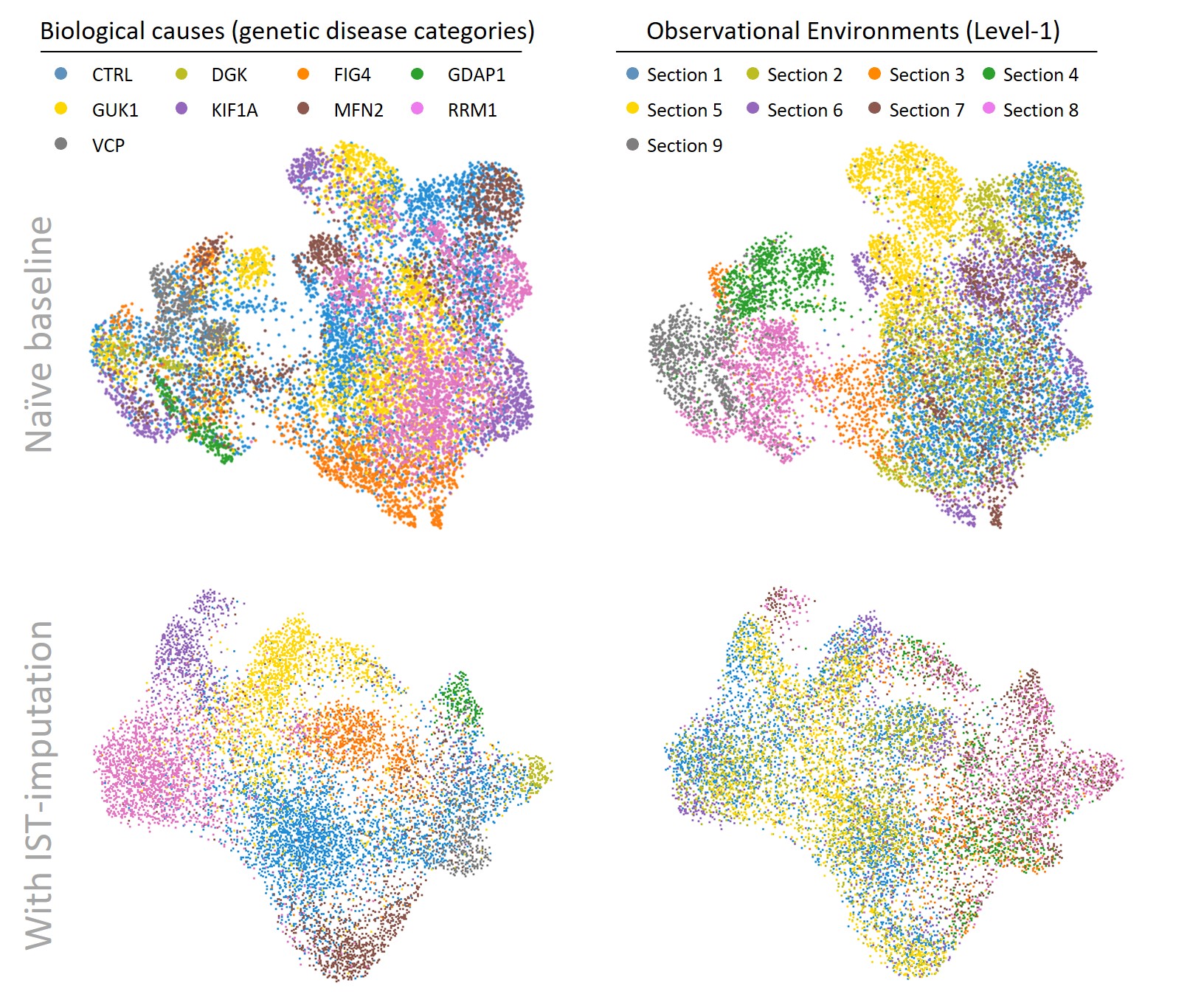}
  \vspace{-4mm}
  \caption{\small{UMAPs of GRID-data training-set features extracted from the penultimate layer of predictors trained with or without IST. Colors show disease-categories $Y$ (left) and OEs $C$ (right)}}
  \label{fig:Figure_4_UMAPs}
  \vspace{-4mm}
\end{figure}

\noindent \textbf{Domain-Adversarial Regularization:} We also compare to domain-adversarial (DR) training as a a regularization technique to learn features that discriminate classes but are invariant to domain-shifts between datasets \cite{ganin2016domain}. We adopt this strategy with slight modification to allow for multiple domains (OEs). Specifically, we modify the architecture of our naive baseline by adding a second classification-head that distinguishes OEs. During backpropagation, we employ a gradient-reversal layer to invert the gradient emanating from the OE-classifier for all layers in the shared ResNet18 stem. We tuned DR's gradient-reversal hyperparamter $\lambda$ by grid-search to optimize validation accuracy on $Y$, while minimizing performance on $C$. \vadjust{\vspace{3pt}}

\begin{table*}
  \centering
  \scriptsize{
  \begin{tabular}{@{}lccc|ccc|cc|ccc@{}}
    \toprule
                    \multicolumn{8}{c}{\textbf{GRID}} &\multicolumn{2}{c}{\textbf{LINCS-SC}} \\
    \cmidrule(lr){2-7}  \cmidrule(lr){8-12}              
                    & IID   & cLISI/bLISI   & kNN-CV   & Level-1   & Level-2   & Level-3   & IID   & kNN-CV   & Level-1   & Level-2   & Level-3   \\
    \midrule
     Baseline   & 0.55  & 0.5417 & 0.4458    & 0.1877    & 0.1381    & 0.1254    & \textbf{0.57} & 0.1271    & NA   & 0.4194  & 0.0405    \\
     Baseline (kNN)   & 0.63  & 0.5417    & 0.4458    & 0.1838    & 0.1378    & 0.1317    & 0.55  & 0.1271    & NA & 0.3897 & 0.0452   \\
     Symphony (kNN)   & 0.50  & 0.7340    & 0.4404    & 0.1797    & 0.1474    & 0.1325    & 0.35  & 0.1098    & NA & 0.2697 & 0.0461 \\
     DR $\alpha=0.0625$ & \textbf{0.73} & 1.0811    & \textbf{0.6906}   & 0.1900    & 0.1379    & 0.1259   & 0.56  & 0.1381    & NA  & 0.4256  & 0.0438  \\
     MixStyle-DA    & 0.60    & 0.6039    & 0.5519   & 0.1084    & nc    & nc   & 0.41  & nc   & NA  & 0.4250  & 0.0450    \\
     StarGANv2 & 0.20   & 1.099 & 0.1539   & 0.1659    & 0.1284    & 0.0977   & nc  & nc    & NA  & nc  & nc    \\
     IST (ours) & 0.60  & \textbf{1.4963}    & 0.5815    & \textbf{0.5839}   & \textbf{0.5350}  & \textbf{0.3673} & 0.53  & \textbf{0.3304} & NA  & \textbf{0.7016}   & \textbf{0.3138}   \\
    \bottomrule
  \end{tabular}}
  \caption{Macro f1 and LISI scores on predictor performance on GRID and LINCS-SC. We report kNN-based classification results for Symphony vs. baseline. For all, higher is better. Level-1 cannot be evaluated for LINCS-SC (see Fig. \ref{fig:Figure_2_Dataset_Overview}C). nc: not computed.}
  \label{tab:results_table}
\end{table*}

\noindent \textbf{StarGANv2:} We assess StarGANv2 (SG2) \cite{choi2020stargan} as a SOTA style-transfer method in natural images. We train using default parameters over 75k iterations. We sample content and style image pairs as for IST and use OE-labels $\mathbf{Y}$ as domains in SG2's multi-task discriminator. Following training, we use SG2 in the same way as IST, to project input images to random OEs, in the hope to yield an interventional training distribution free of spurious correlations. \vadjust{\vspace{3pt}}

\noindent \textbf{MixStyle:} Finally, we assess MixStyle (MS) \cite{zhou2021domain} as a second recent style-transfer baseline that was specifically developed for domain generalization. We implement MS in our predictor architecture and successfully recapitulate \cite{zhou2021domain}'s results on PACS using our setup. For fairer comparison to IST, we also test MS in a domain adaptation setup, in which we allow MS to train on styles (but not biological causes $\mathbf{Y}$) of images from test OEs. \vadjust{\vspace{3pt}}

\subsection{Evaluation metrics}

\noindent \textbf{OOD-generalization:} To test OOD-generalization, we perform section-, plate-, or series-wise cross-validation (\emph{levels of generalization}, see Fig. \ref{fig:Figure_2_Dataset_Overview}) by testing predictors on OEs that were left out during training. \vadjust{\vspace{3pt}}

\noindent \textbf{UMAPs:} While qualitative, feature-space visualizations are widely used in the biomedical literature. We report Uniform Manifold Approximation and Projection (UMAPs) \cite{mcinnes2018umap}. \vadjust{\vspace{3pt}}

\noindent \textbf{LISI score:} We use a ratio of OE- (bLISI) to cause-wise (cLISI) scores \cite{korsunsky2019fast}, normalized over the cardinality $|C|$ and $|Y|$ respectively, to quantify local diversity in feature space. Ideally, bLISI $=1$, while cLISI $=1/|Y|$. \vadjust{\vspace{3pt}}

\noindent \textbf{kNN-CV:} As a second feature-space based metric, we simulate our OOD-generalization experiments by evaluating kNN-classifiers on predicting cause-labels for validation-set images from OEs the corresponding training set images of which are left out of the kNN-reference set. \vadjust{\vspace{3pt}}

\section{Results} 
\label{sec:Results}
\noindent We report empirical results for IST and all baselines for GRID and validate our results on LINCS-SC (Table \ref{tab:results_table}). Trained across all OEs, our naive baseline achieves excellent performance on IID hold-out data, suggesting there exist robust phenotypic manifestations of inherent genetic (GRID) and pharmacological (LINCS-SC) causes in the observed single-cell images. However, visual inspection of the resulting feature spaces via UMAP reveals OEs as a prominent superstructure in our models' representations, whereas biological causes form secondary clusters within the local context of their parent OE (Fig. \ref{fig:Figure_4_UMAPs}). Consistently, LISI scores indicate poor integration over OEs and performance deteriorates in kNN-CV. Critically, when tested on OOD-generalization, our naive baseline shows almost complete collapse across all three levels of generalization on GRID and level-3 for LINCS-SC. \vadjust{\vspace{-1pt}}

 As expected, we find that SYM excels at purging variation over OEs when assessing LISI-scores on the training set. However, for both GRID and LINCS-SC data, we find that this effect does not generalize even to IID validation data and performs poorly on all other metrics. We find that DR-models achieve LISI-scores similar to SYM on GRID, while fully generalizing to IID data, and drastically improving kNN-CV scores. On LINCS-SC however, DR yields only comparatively minor improvements in kNN-CV scores. Remarkably, despite these somewhat promising auxiliary metrics, DR does not significantly improve OOD-generalization across any level in either dataset. Likely because SG2 permutes both style and content (see Fig. \ref{fig:Figure_2_SG2}), predictors trained on the SG2-generated distribution fail even at IID generalization. MixStyle on the other hand, yields excellent IID-performances but - presumably hampered by it's assumptions about what constitutes style-features - yields equally disappointing results in our OOD tests.  \vadjust{\vspace{-1pt}}

By contrast, we find that IST learns to faithfully impute observations as if they had been made in different OEs (Fig. \ref{fig:Figure_3_IST_MainFigure}B). Qualitative inspection of output images suggests that IST simultaneously preserves phenotypic content of the source images (Fig.\ref{fig:Figure_3_IST_MainFigure}C). As such, IST is able to randomize over the confounder $C$ (Fig.\ref{fig:Figure_3_IST_MainFigure}B) to yield a training distribution $P(y,\hat{x},z|do(c))$ in which the original correlations between OEs and biological causes are diminished. Consistent with this, we observe major performance gains across all levels of OOD-generalization, as well as other metrics, for both GRID and LINCS-SC data, when predictors are trained on IST-generated data-distributions (Table \ref{tab:results_table}). These results suggest that our IST approach generates effective interventions on confounders and thereby promotes the emergence of causal representations of biological phenomena.

\section{Conclusions}
\label{sec:Conclusions}

\noindent Learning visual features that generalize across environments is a critical prerequisite for real-world applications of machine learning systems in biomedicine, yet the field lacks broadly adopted metrics to assess progress towards this goal. We propose OOD-generalization tests structured according to a hierarchy of technical processing steps that generally characterize the data generation process of most high-content imaging studies. We show that seemingly well-performing baselines, including SOTA-methods for batch-effect correction, as assessed by IID hold-out sets and several auxiliary metrics, almost completely collapse on this benchmark, revealing highly confounded representations. The success of IST instead shows that effective interventions to mitigate confounders can be learned, given they are at least partially observed. We point out that even models trained on billions of diverse natural images have only achieved minor gains on ObjectNet \cite{goyal2022vision}, suggesting that scale alone is not efficient at breaking contextual biases. Conversely, we suggest our approach bears semblance to thought experiments, by which humans routinely reevaluate familiar concepts in never-observed contexts, thus filling in a sparse matrix of actual observations. We propose IST as a fruitful direction for efficient causal learning. 

\subsection{Acknowledgements} This research is based on work partially supported by Muscular Dystrophy Association (MDA) Development Grant 628114, and NIH award K99HG011488 to WMP, and by the Broad Institute Schmidt Fellowship program (JCC) and by NSF award 2134695 to JCC.

\section*{Competing interests}
\noindent Columbia University has filed a US patent application on Interventional Style Transfer for causal representation learning on behalf of WMP, JCC and MH. The remaining authors declare no competing interests.

{\small
\bibliographystyle{ieee_fullname}
\bibliography{references}

\begin{thebibliography}{10}\itemsep=-1pt

\bibitem{arjovsky2019invariant}
Martin Arjovsky, L{\'e}on Bottou, Ishaan Gulrajani, and David Lopez-Paz.
\newblock Invariant risk minimization.
\newblock {\em arXiv preprint arXiv:1907.02893}, 2019.

\bibitem{barbu2019objectnet}
Andrei Barbu, David Mayo, Julian Alverio, William Luo, Christopher Wang, Dan
  Gutfreund, Josh Tenenbaum, and Boris Katz.
\newblock Objectnet: A large-scale bias-controlled dataset for pushing the
  limits of object recognition models.
\newblock {\em Advances in neural information processing systems}, 32, 2019.

\bibitem{bray2016cell}
Mark-Anthony Bray, Shantanu Singh, Han Han, Chadwick~T Davis, Blake Borgeson,
  Cathy Hartland, Maria Kost-Alimova, Sigrun~M Gustafsdottir, Christopher~C
  Gibson, and Anne~E Carpenter.
\newblock Cell painting, a high-content image-based assay for morphological
  profiling using multiplexed fluorescent dyes.
\newblock {\em Nature protocols}, 11(9):1757--1774, 2016.

\bibitem{caron2020unsupervised}
Mathilde Caron, Ishan Misra, Julien Mairal, Priya Goyal, Piotr Bojanowski, and
  Armand Joulin.
\newblock Unsupervised learning of visual features by contrasting cluster
  assignments.
\newblock {\em Advances in Neural Information Processing Systems},
  33:9912--9924, 2020.

\bibitem{caron2021emerging}
Mathilde Caron, Hugo Touvron, Ishan Misra, Herv{\'e} J{\'e}gou, Julien Mairal,
  Piotr Bojanowski, and Armand Joulin.
\newblock Emerging properties in self-supervised vision transformers.
\newblock In {\em Proceedings of the IEEE/CVF International Conference on
  Computer Vision}, pages 9650--9660, 2021.

\bibitem{chandrasekaran2023jump}
Srinivas~Niranj Chandrasekaran, Jeanelle Ackerman, Eric Alix, D~Michael Ando,
  John Arevalo, Melissa Bennion, Nicolas Boisseau, Adriana Borowa, Justin~D
  Boyd, Laurent Brino, et~al.
\newblock Jump cell painting dataset: morphological impact of 136,000 chemical
  and genetic perturbations.
\newblock {\em bioRxiv}, pages 2023--03, 2023.

\bibitem{chen2020simple}
Ting Chen, Simon Kornblith, Mohammad Norouzi, and Geoffrey Hinton.
\newblock A simple framework for contrastive learning of visual
  representations.
\newblock In {\em International conference on machine learning}, pages
  1597--1607. PMLR, 2020.

\bibitem{choi2017stargan}
Yunjey Choi, Min-Je Choi, Munyoung Kim, Jung-Woo Ha, Sunghun Kim, and Jaegul
  Choo.
\newblock Stargan: unified generative adversarial networks for multi-domain
  image-to-image translation. corr abs/1711.09020 (2017).
\newblock {\em arXiv preprint arXiv:1711.09020}, 2017.

\bibitem{choi2020stargan}
Yunjey Choi, Youngjung Uh, Jaejun Yoo, and Jung-Woo Ha.
\newblock Stargan v2: Diverse image synthesis for multiple domains.
\newblock In {\em Proceedings of the IEEE/CVF conference on computer vision and
  pattern recognition}, pages 8188--8197, 2020.

\bibitem{corsello2017drug}
Steven~M Corsello, Joshua~A Bittker, Zihan Liu, Joshua Gould, Patrick McCarren,
  Jodi~E Hirschman, Stephen~E Johnston, Anita Vrcic, Bang Wong, Mariya Khan,
  et~al.
\newblock The drug repurposing hub: a next-generation drug library and
  information resource.
\newblock {\em Nature medicine}, 23(4):405--408, 2017.

\bibitem{zhou2021domain}
Zhou et. al.
\newblock Domain generalization with mixstyle.
\newblock {\em ICLR 2021}.

\bibitem{fay2023rxrx3}
Marta~M Fay, Oren Kraus, Mason Victors, Lakshmanan Arumugam, Kamal Vuggumudi,
  John Urbanik, Kyle Hansen, Safiye Celik, Nico Cernek, Ganesh Jagannathan,
  et~al.
\newblock Rxrx3: Phenomics map of biology.
\newblock {\em bioRxiv}, pages 2023--02, 2023.

\bibitem{ganin2016domain}
Yaroslav Ganin, Evgeniya Ustinova, Hana Ajakan, Pascal Germain, Hugo
  Larochelle, Fran{\c{c}}ois Laviolette, Mario Marchand, and Victor Lempitsky.
\newblock Domain-adversarial training of neural networks.
\newblock {\em The journal of machine learning research}, 17(1):2096--2030,
  2016.

\bibitem{geirhos2018imagenet}
Robert Geirhos, Patricia Rubisch, Claudio Michaelis, Matthias Bethge, Felix~A
  Wichmann, and Wieland Brendel.
\newblock Imagenet-trained cnns are biased towards texture; increasing shape
  bias improves accuracy and robustness.
\newblock {\em arXiv preprint arXiv:1811.12231}, 2018.

\bibitem{georgopoulos2021mitigating}
Markos Georgopoulos, James Oldfield, Mihalis~A Nicolaou, Yannis Panagakis, and
  Maja Pantic.
\newblock Mitigating demographic bias in facial datasets with style-based
  multi-attribute transfer.
\newblock {\em International Journal of Computer Vision}, 129(7):2288--2307,
  2021.

\bibitem{goldsborough2017cytogan}
Peter Goldsborough, Nick Pawlowski, Juan~C Caicedo, Shantanu Singh, and Anne~E
  Carpenter.
\newblock Cytogan: generative modeling of cell images.
\newblock {\em BioRxiv}, page 227645, 2017.

\bibitem{goyal2022vision}
Priya Goyal, Quentin Duval, Isaac Seessel, Mathilde Caron, Mannat Singh, Ishan
  Misra, Levent Sagun, Armand Joulin, and Piotr Bojanowski.
\newblock Vision models are more robust and fair when pretrained on uncurated
  images without supervision.
\newblock {\em arXiv preprint arXiv:2202.08360}, 2022.

\bibitem{he2016deep}
Kaiming He, Xiangyu Zhang, Shaoqing Ren, and Jian Sun.
\newblock Deep residual learning for image recognition.
\newblock In {\em Proceedings of the IEEE conference on computer vision and
  pattern recognition}, pages 770--778, 2016.

\bibitem{huang2017arbitrary}
Xun Huang and Serge Belongie.
\newblock Arbitrary style transfer in real-time with adaptive instance
  normalization.
\newblock In {\em Proceedings of the IEEE international conference on computer
  vision}, pages 1501--1510, 2017.

\bibitem{isola2017image}
Phillip Isola, Jun-Yan Zhu, Tinghui Zhou, and Alexei~A Efros.
\newblock Image-to-image translation with conditional adversarial networks.
\newblock In {\em Proceedings of the IEEE conference on computer vision and
  pattern recognition}, pages 1125--1134, 2017.

\bibitem{kang2021efficient}
Joyce~B Kang, Aparna Nathan, Kathryn Weinand, Fan Zhang, Nghia Millard, Laurie
  Rumker, D Moody, Ilya Korsunsky, and Soumya Raychaudhuri.
\newblock Efficient and precise single-cell reference atlas mapping with
  symphony.
\newblock {\em Nature communications}, 12(1):1--21, 2021.

\bibitem{karras2020analyzing}
Tero Karras, Samuli Laine, Miika Aittala, Janne Hellsten, Jaakko Lehtinen, and
  Timo Aila.
\newblock Analyzing and improving the image quality of stylegan.
\newblock In {\em Proceedings of the IEEE/CVF conference on computer vision and
  pattern recognition}, pages 8110--8119, 2020.

\bibitem{korsunsky2019fast}
Ilya Korsunsky, Nghia Millard, Jean Fan, Kamil Slowikowski, Fan Zhang, Kevin
  Wei, Yuriy Baglaenko, Michael Brenner, Po-ru Loh, and Soumya Raychaudhuri.
\newblock Fast, sensitive and accurate integration of single-cell data with
  harmony.
\newblock {\em Nature methods}, 16(12):1289--1296, 2019.

\bibitem{lang2021explaining}
Oran Lang, Yossi Gandelsman, Michal Yarom, Yoav Wald, Gal Elidan, Avinatan
  Hassidim, William~T Freeman, Phillip Isola, Amir Globerson, Michal Irani,
  et~al.
\newblock Explaining in style: Training a gan to explain a classifier in
  stylespace.
\newblock In {\em Proceedings of the IEEE/CVF International Conference on
  Computer Vision}, pages 693--702, 2021.

\bibitem{lee2009advances}
Daniel~D Lee, P Pham, Y Largman, and A Ng.
\newblock Advances in neural information processing systems 22.
\newblock Technical report, Tech. Rep., Tech. Rep, 2009.

\bibitem{li2017universal}
Yijun Li, Chen Fang, Jimei Yang, Zhaowen Wang, Xin Lu, and Ming-Hsuan Yang.
\newblock Universal style transfer via feature transforms.
\newblock {\em Advances in neural information processing systems}, 30, 2017.

\bibitem{li2017demystifying}
Yanghao Li, Naiyan Wang, Jiaying Liu, and Xiaodi Hou.
\newblock Demystifying neural style transfer.
\newblock {\em arXiv preprint arXiv:1701.01036}, 2017.

\bibitem{mao2021generative}
Chengzhi Mao, Augustine Cha, Amogh Gupta, Hao Wang, Junfeng Yang, and Carl
  Vondrick.
\newblock Generative interventions for causal learning.
\newblock In {\em Proceedings of the IEEE/CVF Conference on Computer Vision and
  Pattern Recognition}, pages 3947--3956, 2021.

\bibitem{mcinnes2018umap}
Leland McInnes, John Healy, and James Melville.
\newblock Umap: Uniform manifold approximation and projection for dimension
  reduction.
\newblock {\em arXiv preprint arXiv:1802.03426}, 2018.

\bibitem{mcquin2018cellprofiler}
Claire McQuin, Allen Goodman, Vasiliy Chernyshev, Lee Kamentsky, Beth~A Cimini,
  Kyle~W Karhohs, Minh Doan, Liya Ding, Susanne~M Rafelski, Derek Thirstrup,
  et~al.
\newblock Cellprofiler 3.0: Next-generation image processing for biology.
\newblock {\em PLoS biology}, 16(7):e2005970, 2018.

\bibitem{moshkov2022learning}
Nikita Moshkov, Michael Bornholdt, Santiago Benoit, Claire McQuin, Matthew
  Smith, Allen Goodman, Rebecca Senft, Yu Han, Mehrtash Babadi, Peter Horvath,
  et~al.
\newblock Learning representations for image-based profiling of perturbations.
\newblock {\em bioRxiv}, 2022.

\bibitem{mullard2019machine}
Asher Mullard.
\newblock Machine learning brings cell imaging promises into focus.
\newblock {\em Nat. Rev. Drug Discovery 2019}.

\bibitem{pearl2009causality}
Judea Pearl.
\newblock {\em Causality}.
\newblock Cambridge university press, 2009.

\bibitem{qian2020batch}
Wesley~Wei Qian, Cassandra Xia, Subhashini Venugopalan, Arunachalam
  Narayanaswamy, Michelle Dimon, George~W Ashdown, Jake Baum, Jian Peng, and
  D~Michael Ando.
\newblock Batch equalization with a generative adversarial network.
\newblock {\em Bioinformatics}, 36(Supplement\_2):i875--i883, 2020.

\bibitem{recht2019imagenet}
Benjamin Recht, Rebecca Roelofs, Ludwig Schmidt, and Vaishaal Shankar.
\newblock Do imagenet classifiers generalize to imagenet?
\newblock In {\em International Conference on Machine Learning}, pages
  5389--5400. PMLR, 2019.

\bibitem{rombach2022high}
Robin Rombach, Andreas Blattmann, Dominik Lorenz, Patrick Esser, and Bj{\"o}rn
  Ommer.
\newblock High-resolution image synthesis with latent diffusion models.
\newblock In {\em Proceedings of the IEEE/CVF Conference on Computer Vision and
  Pattern Recognition}, pages 10684--10695, 2022.

\bibitem{ronneberger2015u}
Olaf Ronneberger, Philipp Fischer, and Thomas Brox.
\newblock U-net: Convolutional networks for biomedical image segmentation.
\newblock In {\em International Conference on Medical image computing and
  computer-assisted intervention}, pages 234--241. Springer, 2015.

\bibitem{schiff2022integrating}
Lauren Schiff, Bianca Migliori, Ye Chen, Deidre Carter, Caitlyn Bonilla, Jenna
  Hall, Minjie Fan, Edmund Tam, Sara Ahadi, Brodie Fischbacher, et~al.
\newblock Integrating deep learning and unbiased automated high-content
  screening to identify complex disease signatures in human fibroblasts.
\newblock {\em Nature Communications}, 13(1):1590, 2022.

\bibitem{scholkopf2022causality}
Bernhard Sch{\"o}lkopf.
\newblock Causality for machine learning.
\newblock In {\em Probabilistic and Causal Inference: The Works of Judea
  Pearl}, pages 765--804. 2022.

\bibitem{scholkopf2021toward}
Bernhard Sch{\"o}lkopf, Francesco Locatello, Stefan Bauer, Nan~Rosemary Ke, Nal
  Kalchbrenner, Anirudh Goyal, and Yoshua Bengio.
\newblock Toward causal representation learning.
\newblock {\em Proceedings of the IEEE}, 109(5):612--634, 2021.

\bibitem{tian2020rethinking}
Yonglong Tian, Yue Wang, Dilip Krishnan, Joshua~B Tenenbaum, and Phillip Isola.
\newblock Rethinking few-shot image classification: a good embedding is all you
  need?
\newblock In {\em European Conference on Computer Vision}, pages 266--282.
  Springer, 2020.

\bibitem{wang2017effectiveness}
Jason Wang, Luis Perez, et~al.
\newblock The effectiveness of data augmentation in image classification using
  deep learning.
\newblock {\em Convolutional Neural Networks Vis. Recognit}, 11:1--8, 2017.

\bibitem{way2022morphology}
Gregory~P Way, Ted Natoli, Adeniyi Adeboye, Lev Litichevskiy, Andrew Yang,
  Xiaodong Lu, Juan~C Caicedo, Beth~A Cimini, Kyle Karhohs, David~J Logan,
  et~al.
\newblock Morphology and gene expression profiling provide complementary
  information for mapping cell state.
\newblock {\em Cell Systems}, 2022.

\bibitem{xie2020adversarial}
Cihang Xie, Mingxing Tan, Boqing Gong, Jiang Wang, Alan~L Yuille, and Quoc~V
  Le.
\newblock Adversarial examples improve image recognition.
\newblock In {\em Proceedings of the IEEE/CVF Conference on Computer Vision and
  Pattern Recognition}, pages 819--828, 2020.

\bibitem{yang2021mol2image}
Karren Yang, Samuel Goldman, Wengong Jin, Alex~X Lu, Regina Barzilay, Tommi
  Jaakkola, and Caroline Uhler.
\newblock Mol2image: Improved conditional flow models for molecule to image
  synthesis.
\newblock In {\em Proceedings of the IEEE/CVF Conference on Computer Vision and
  Pattern Recognition}, pages 6688--6698, 2021.

\bibitem{yucer2020exploring}
Seyma Yucer, Samet Ak{\c{c}}ay, Noura Al-Moubayed, and Toby~P Breckon.
\newblock Exploring racial bias within face recognition via per-subject
  adversarially-enabled data augmentation.
\newblock In {\em Proceedings of the IEEE/CVF Conference on Computer Vision and
  Pattern Recognition Workshops}, pages 18--19, 2020.

\bibitem{zhu2017unpaired}
Jun-Yan Zhu, Taesung Park, Phillip Isola, and Alexei~A Efros.
\newblock Unpaired image-to-image translation using cycle-consistent
  adversarial networks.
\newblock In {\em Proceedings of the IEEE international conference on computer
  vision}, pages 2223--2232, 2017.

\end{thebibliography}
}

\clearpage

\beginsupplement
\strut
\FloatBarrier
\section{Supplementary Material}
\subsection{Extended technical analysis and limitations:}
\label{sec:ExtendedTech}
\subsubsection{Ablations:} 
\begin{figure*}
  \centering
  \includegraphics[width=0.90\linewidth]{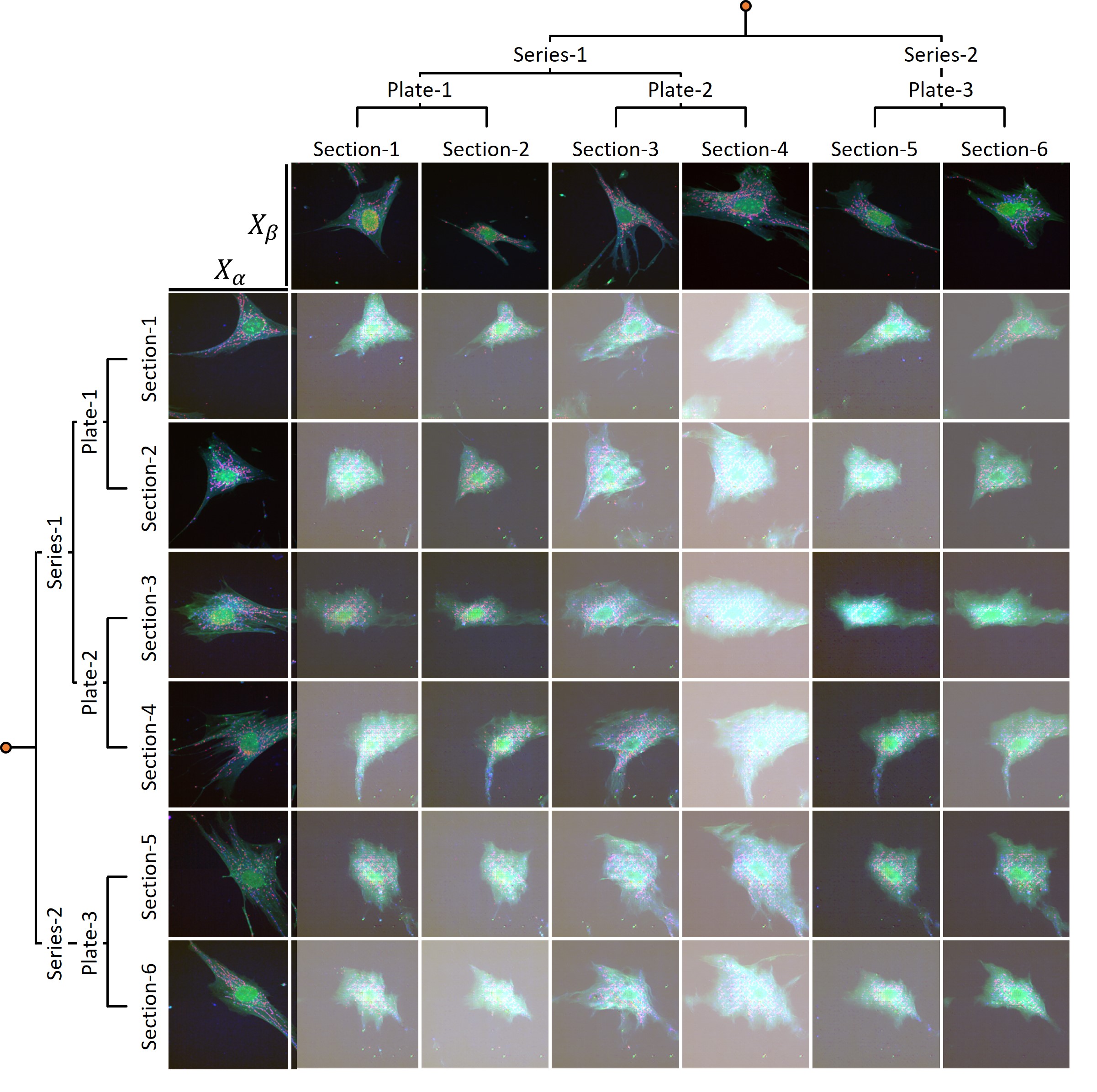}
  \vspace{-4mm}
  \caption{\small{Substantially degraded IST-output image quality for an IST model with ablated UNet skip-connections. Images $X_\alpha$ and $X_\beta$ are the same as in \ref{fig:Figure_3_IST_MainFigure}B.}}
  \label{fig:GRID_images_noUNet}
  \vspace{-4mm}
\end{figure*}

To provide further insight into key components of our method, we report a limited ablation study, evaluating their effects on IST's ability to improve OOD generalization against a single level-1 fold of GRID (Table \ref{tab:GRID_ablations_rxiv}). Ablation of individual loss terms highlights $\mathcal{L}_{Cmatch}$ and $\mathcal{L}_{Style}$ as key components of our full objective, whereas ablation of $\mathcal{L}_{Cycle}$ has only a minor impact on performance. Removing $\mathcal{L}_{Cont}$ improved performance by 2 points. While this may suggest that discouraging alterations in pixel space through $\mathcal{L}_{Cont}$ may be too restrictive in this case, we prefer to err on the side of caution, wary that the preservation of (phenotypic) content is vital to any applications of IST in scientific discovery (Combine ablations on "content" terms to check for redundancy). A key architectural prior we introduce to this end are UNet skip-connections between the encoder and decoder branches of our IST generator (see Sec. \ref{sec:Method}, Fig. \ref{fig:Figure_3_IST_MainFigure}) \cite{ronneberger2015u}. We show that training our IST-model without UNet skip-connections greatly deteriorates the quality of the resulting IST output images (Fig. \ref{fig:GRID_images_noUNet}).

We also investigate the extent to which the performance of our method depends on the number and diversity of style-target images available during inference. Particularly worrisome would be any evidence of content leakage from style-target images. Although the strategy by which we sample style-targets during training, i.e. randomly balanced over both causes (class labels) and OEs (see Sec. \ref{sec:Method}), is designed to avoid this, it could be insufficient. Instead, we find that reducing the diversity of available style-targets during inference to even a single style-target image per OE, irrespective of their class labels, if anything, slightly improves the ability of IST models to produce effective interventional training distributions (Table \ref{tab:GRID_ablations_rxiv}). 

Finally, we provide empirical evidence for the benefit of projecting source images into the IST-reference space for the performance of predictors trained on IST-generated interventional distributions, as done by default in our experiments (see Sec. \ref{sec:Method}). We find that test time performance of IST-trained predictors deteriorates substantially when evaluated directly on test images, although even here, IST-predictors retain a large performance gain over other baselines. We leave further interpretations to future work.  

\begin{table}
  \centering
  \scriptsize{
  \begin{tabular}{@{}lc|c|c@{}}
    \toprule
                    & IID   & Level-1 (vs Section 6)  & Level-1  \\
    \midrule
     1. Baseline (ERM)       & 0.55  & 0.1965 \textcolor{red}{(-0.521)} & 0.1877    \\
     2. \textbf{IST (ours)}     & 0.6013  & \textbf{0.7176} (+0.000) & \textbf{0.5839}   \\
    \midrule
     3. IST w/o $\mathcal{L}_{Cmatch}$       & 0.3961  & 0.2706 \textcolor{red}{(-0.447)}  & -               \\
     4. IST w/o $\mathcal{L}_{Style}$     & 0.6972  & 0.5885  \textcolor{red}{(-0.122)} & -                 \\
     5. IST w/o $\mathcal{L}_{Cycle}$      & 0.6426  & 0.7053 \textcolor{red}{(-0.012)} & -                 \\
     6. IST w/o $\mathcal{L}_{Cont}$      & 0.6149  & 0.7449 \textcolor{green}{(+0.027)}  & -                 \\
    \midrule
     7. IST one-Img/OE      & 0.6061  & 0.7315 \textcolor{green}{(-0.014)} & -                 \\
     8. Ours w/o test-time adapt.      & 0.6046    & 0.5042 \textcolor{red}{(-0.213)}    & - \\

    \bottomrule
  \end{tabular}}
  \caption{Macro f1-scores for additional baselines \& ablations on GRID. A single (random) level-1 fold was assessed for ablations.}
  \label{tab:GRID_ablations_rxiv}
\end{table}

\subsubsection{Limitations:}
Despite it's success, our IST method has limitations and - as the preceding section suggests - offers ample room for improvement. Given the rapid technical progress in the field, our implementation of the IST generator is almost certainly sub-optimal, and recent advancements in image-generation (e.g. \cite{rombach2022high}) are likely to yield large gains in image-realism and IST fidelity. We also note that our IST-implementation incorporates several priors that help define content in our data types by virtue of discouraging substantial changes in pixel space (see Sec. \ref{sec:Method}). While the value of these priors are - as we show - borne out empirically for microscopy data, and while we expect that this will hold true for biomedical data broadly, these priors are likely incompatible with (natural) image types in which e.g. 3D rotations or changing viewpoint angles constitute a substantial factor of variation across observational environments. Finally, the performance of our system currently benefits substantially from the use of class-labels during training (see Table \ref{tab:GRID_ablations_rxiv}.3). This is undesirable, even if labels are, as is common in biomedical studies, implicit to the data generation process (we know the perturbations). Advances in self-supervised learning \cite{caron2021emerging}\cite{caron2020unsupervised}\cite{chen2020simple} may allow future work to avoid this dependency. 

\subsection{Synthetic OOD-experiments}
The rationale behind our IST-approach is guided by the idea that models encode spurious correlations between confounding nuisances and causes in the training data and consequently fail to generalize to OOD data. How OEs as confounders manifest in the complex single-cell fluorescent microscopy images we study in this work however is not obvious even to human expert analysts. Our empirical results in the main text notwithstanding, we hence conduct experiments on an easy-to-interpret synthetic dataset with perfectly known causes, confounders, correlations.

\begin{figure*}
  \centering
  \includegraphics[width=0.90\linewidth]{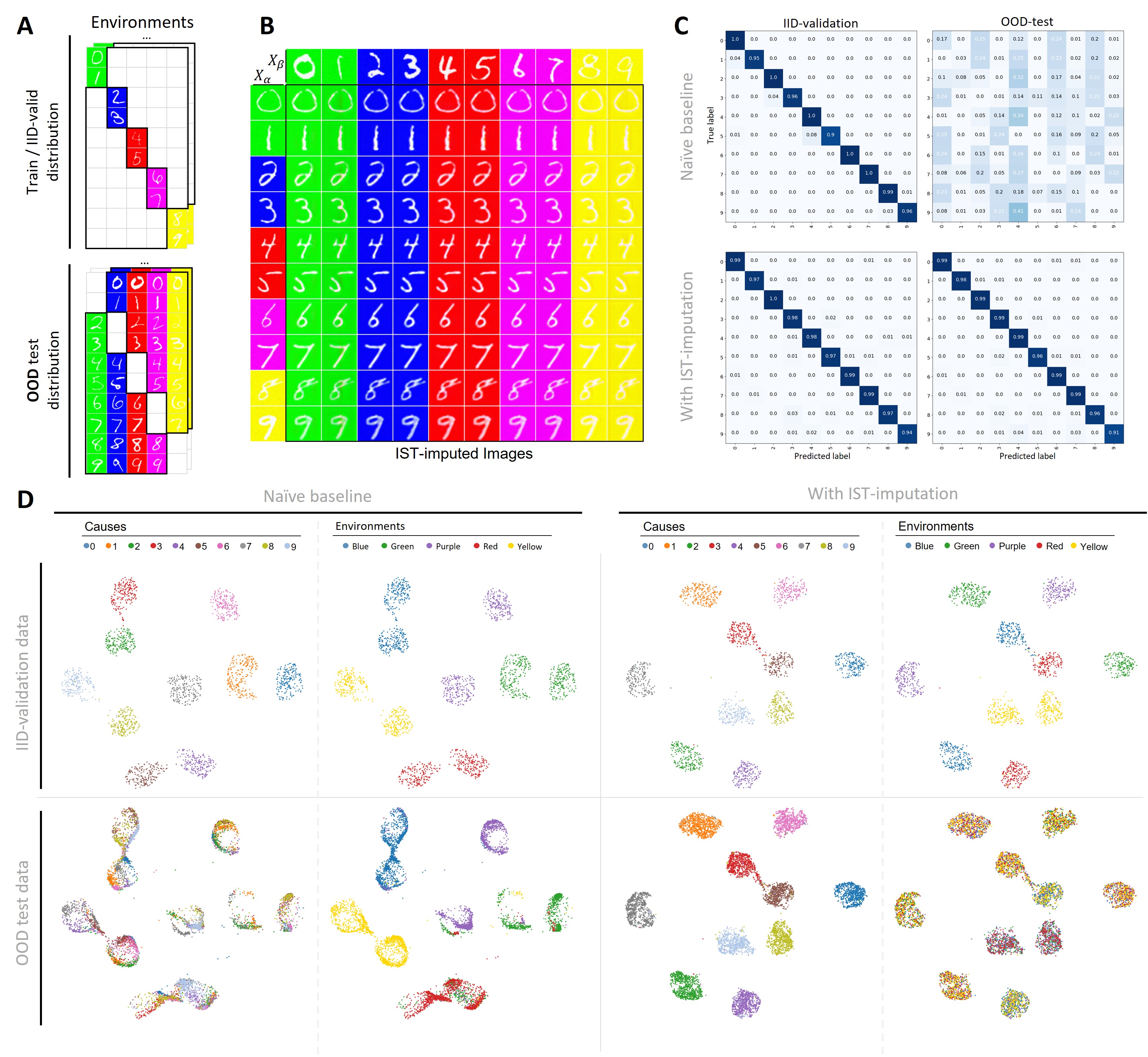}
  \vspace{-4mm}
  \caption{\small{Results on color-MNIST. A: Diagramatic illustration of dataset structure. In the training and IID-validation hold-out sets, pairs of digit-categories only ever encountered with one background color (observational environments), yielding five distinct pairs of digit-categories, and a training distribution that is strongly correlated with confounding nuisances. B: IST-imputed versions of images $x_\alpha$ (sampled from the validation set) based on environment-codes extracted from the indicated images $x_\beta$. C: Confusion matrices. D: UMAP visualizations of the representations $\hat{Z}$ naive baseline or a predictor trained with IST-imputation. UMAPs were computed on training data. Either IID-validation or OOD-test data is shown. Data points in UMAPs are colored are by either cause (digit-category) or observational environment (background color). Note that due to dataset structure, no admixture of environments is expected for IID-validation data for either method.}}
  \label{fig:Supplementary_figure_1}
  \vspace{-4mm}
\end{figure*}

\textbf{Color-MNIST dataset:} We construct a version of color-MNIST \cite{arjovsky2019invariant, mao2021generative} in which digits are confounded by background colors. We use digit categories $y \in \{0,1,..9\}$ as causes and a confounder that introduces spurious correlations between causes and background colors as observational environments. We confound the training and corresponding IID-validation sets by exclusively assigning five background colors $k \in \{green, blue, red, purple, yellow\}$ to mutually exclusive pairs of digit categories. In contrast, the OOD-test set contains all combinations \emph{not} found in the training/validation sets (see Fig. \ref{fig:Supplementary_figure_1}A), thus simulating the extreme scenario in which test data contains causes observed exclusively in OEs different to that of the training set. A predictor that learns to encode the spurious correlations in the training set, is expected to perform well on validation, but poorly (close to chance) on our OOD-test data. If instead causal representations of the digit categories are learned, invariance to background color should allow the predictor to generalize well to both IID and OOD hold-out sets.

\textbf{Experimental Setup:} Contrary to experiments described in the main text, we apply no augmentations or image-level normalizations. Otherwise, we train our naive baseline predictor on digit categories by empirical risk minimization. Next, we train an IST model to impute training-images as if they had been observed in different OEs (i.e. background colors) and leverage the resulting generator to produce an interventional training distribution in which we expect correlations between causes and OEs to be mitigated.

\begin{figure*}
  \centering
  \includegraphics[width=0.90\linewidth]{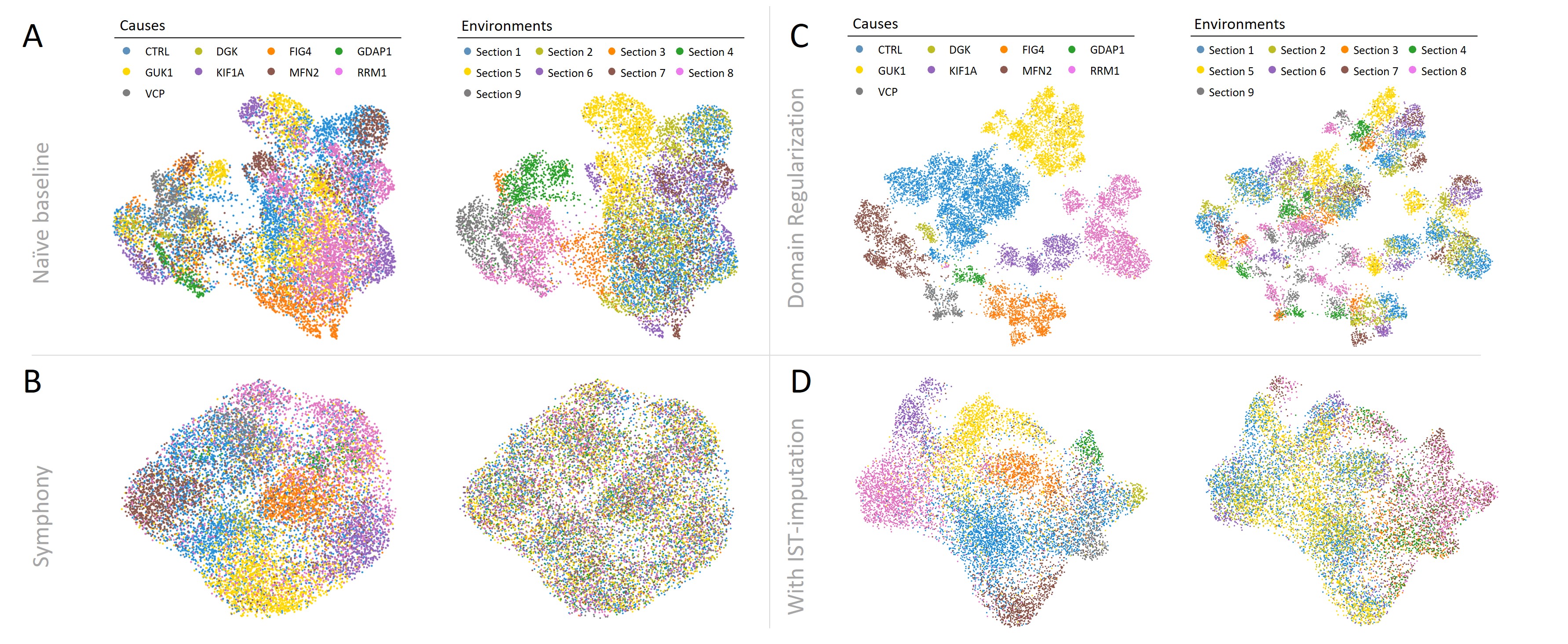}
  \vspace{-4mm}
  \caption{\small{Matrix of UMAPs visualizing causes (disease-categories) $Y$ (left) and series-level environments $C$ (right) for GRID data across all methods.}}
  \label{fig:Suppl_Fig_UMAPs_GRID}
  \vspace{-4mm}
\end{figure*}

\textbf{Empirical results:} As expected, our naive baseline approach achieves excellent performance on hold-out data, as long as the confounding bias inherent to the training dataset is maintained. However, our naive baseline fails to achieve results better than chance on our OOD-test set (Fig. \ref{fig:Supplementary_figure_1}C). Inspection of UMAPs derived from representations $\hat{Z}$ (see \ref{fig:Figure_3_IST_MainFigure}A) illustrate how the naive baseline represents input images exclusively within the landmarks of their corresponding background colors: irrespective of digit-category, we find images with e.g. blue background fall onto the landmarks of digits that were encountered with blue background in the training set (two's and three's) (Fig. \ref{fig:Supplementary_figure_1}D, left). Similar to our results on real-world microscopy data reported in the main text, we note that our model here too appears to directly represents the specific hierarchy underlying our training data, despite not having been trained explicitly to do so.

We find that our trained IST model generates reasonable imputations of source images $X_\alpha$ by virtue of extracted environment (style) codes from some images $X_\beta$ (Fig. \ref{fig:Supplementary_figure_1}B). Moreover, we show that training a predictor based on the IST-distribution $P(X,Y|do(C)$ is effective in yielding representations that - in this case - fully generalize to OOD-test data, achieving performance on par with IID validation data. Consistently, when inspecting the representations $\hat{Z}$ of the resulting IST-predictor model, in UMAP space, there is no observable assortment of images according to background color for OOD-test data, and only by digit category.  

\subsection{Extended data and discussion - GRID:}
We expand on data and discussion included in the main text for our results on GRID. We provide additional UMAP visualizations of $\hat{Z}$ for Symphony and DR baselines on GRID (Fig. \ref{fig:Suppl_Fig_UMAPs_GRID}A-D). For training data, UMAP results for Symphony are especially compelling, with OEs exhibiting close to complete admixture, while class-wise clustering remains largely preserved. However, as shown in Sec.\ref{sec:Results}, we find these effects do not generalize well even to IID hold-out data: despite somewhat improved cLISI/bLISI scores, Symphony deteriorates IID-generalization compared to our naive baseline and fails to improve either kNN-based CV or OOD-generalization scores across any level. 

\begin{figure}
  \centering
  \includegraphics[width=0.90\linewidth]{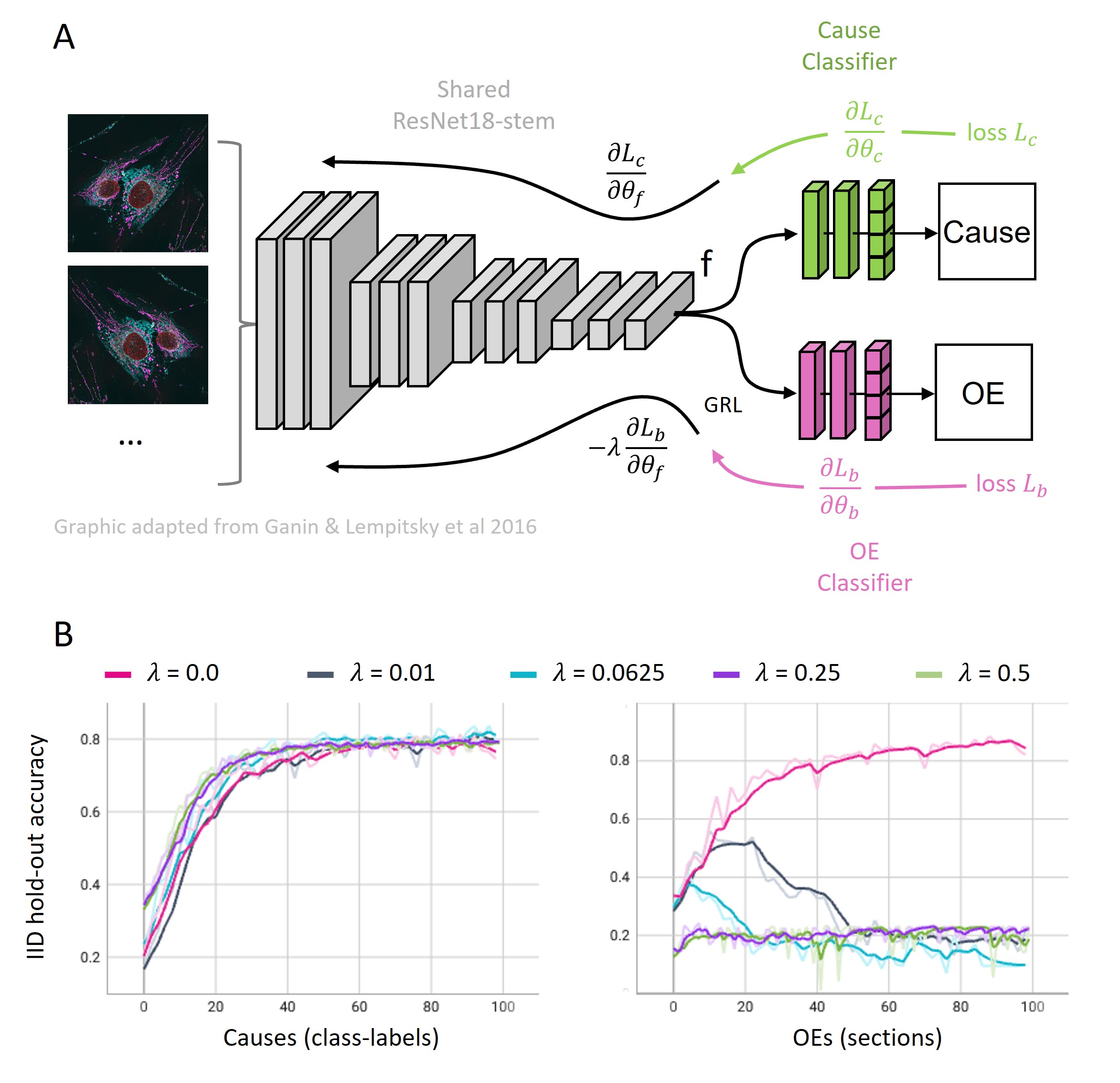}
  \vspace{-4mm}
  \caption{\small{A: Diagrammatic overview over our implementation of DR-regularized training. A Gradient Reversal Layer (GRL) inverts the gradient emmanating from the environment classifier for for the parameters of the shared feature extractor (ResNet18 stem). B: Hyperparameter tuning for DR's $\lambda$ on IID hold-out set accuracy for causes vs. OEs on GRID data}}
  \label{fig:Suppl_Fig_DomainRegularization}
  \vspace{-4mm}
\end{figure}

Consistent with improved LISI scores reported in the main text, we find that DR mitigates the prominence of OEs within UMAP visualizations in favor of biological causes. Still, OEs remain apparent as substructures of causes. This raises the possibility that our choice of the regularization weight $alpha$ (see Fig. \ref{fig:Suppl_Fig_DomainRegularization}A) may have been sub-optimal. We point out however, that we tuned the strength DR's $alpha$ parameter such that DR's OE classification performance remains random (while maximizing performance on causes; Fig. \ref{fig:Suppl_Fig_DomainRegularization}B). Moreover, increasing DR's $alpha$ from 0.0625 to 1.0 has no noticeable effect on UMAP structure (data not shown), which argues against sub-optimal hyperparamter tuning as an explanation for DR's failure to fully remove OEs as a factor of variation learned by predictors. Other avenues, such as increasing capacity of the classification heads, the architecture of which, in the interest of fair comparison to our other baselines, was kept consistent across all experiments, may prove more fruitful. We suspect that the optimal implementation of DR-models in pursuit of improved OOD generalization deserves further attention, but leave such investigations to future work. 

\begin{figure*}
  \centering
  \includegraphics[width=0.90\linewidth]{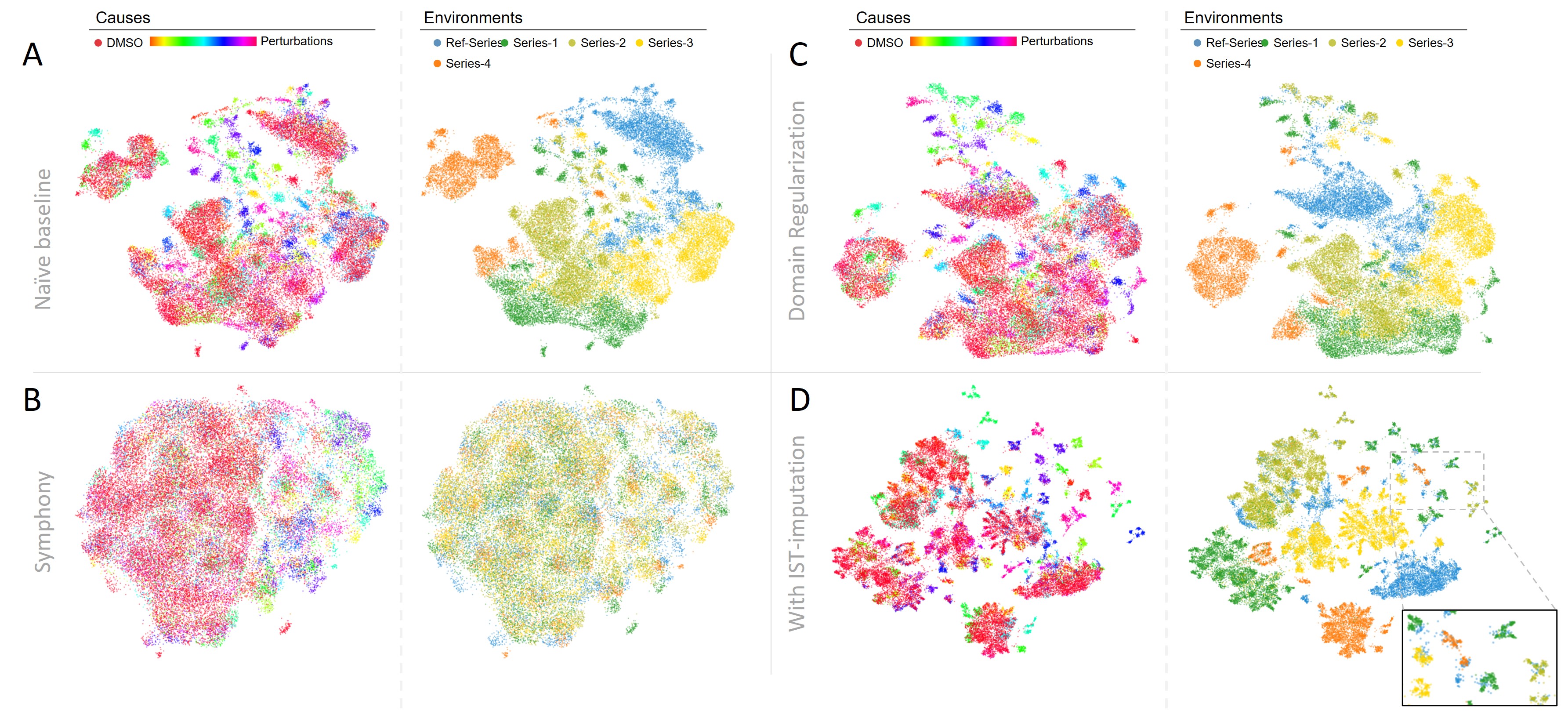}
  \vspace{-4mm}
  \caption{\small{Matrix of UMAPs visualizing causes (perturbations-categories) $Y$ (left) and series-level environments $C$ (right) for LINCS-SC data across all methods. Dimethyl sulfoxide (DMSO) indicates negative control. D: Inset highlights an indicated region of well-resolved data-points clustered by cause, and significant integration with the reference series.}}
  \label{fig:Suppl_Fig_UMAPs_LINCS-SC}
  \vspace{-4mm}
\end{figure*}

\subsection{Extended data and discussion - LINCS-SC:} 
We provide extended data and discussion on our experiments on LINCS-SC. UMAPs of training-set representations of naive baseline models reveal dominant super-structure at the series-level (level-3; Fig. \ref{fig:Suppl_Fig_UMAPs_LINCS-SC}A). Just as for GRID, Symphony achieves strong integration over OEs (Fig. \ref{fig:Suppl_Fig_UMAPs_LINCS-SC}B), however at substantial cost of preserving cause-specific phenotypic signal that identifies a subset of pharmacological perturbations from controls (see Table \ref{tab:results_table} in the main text).

Note that, in contrast to GRID, LINCS-SC contains only a single series that covers all perturbations (Ref-Series), but no replicate plate-maps, while four additional series cover subsets of perturbations over five replicate-plates, but share no perturbations between them (see Fig. \ref{fig:Figure_2_Dataset_Overview}C for reference). Consequently, we would not expect integration between non-reference series, even in perfectly causal representation, unless there was significant overlap between the morphological effects of the perturbations held between them. We note that identifying such points of phenotypic convergence between chemically disparate pharmacological agents is of great interest to the drug-development community, as it might make it possible to infer a new drug-candidates biological \emph{mechanisms of action} (MOA) based on a visual readouts on (sub)cellular morphology. However, the perturbations in LINCS-SC have known MOAs, with little-to-no overlap between them. Hence, we would expect integration only between Ref-Series and each other series, but not among the latter. 

We find that DR-training has no observable impact on diminishing the dominance of series-level (level-3) OEs in the UMAPs. However, consistent with quantitative results, we find substantially increased integration between Ref-Series and other series, when inspecting representations derived from predictors trained with IST (Fig. \ref{fig:Suppl_Fig_UMAPs_LINCS-SC}D).

\end{document}